\documentclass[11pt]{article}

\PassOptionsToPackage{hyperfootnotes=false}{hyperref}
\usepackage[final]{acl}

\usepackage{times}
\usepackage{latexsym}

\usepackage[T1]{fontenc}

\usepackage[utf8]{inputenc}

\usepackage{microtype}

\usepackage{inconsolata}

\usepackage{listings}
\usepackage{graphicx}
\graphicspath{{./}}
\usepackage{subcaption}
\usepackage{multirow}
\usepackage{booktabs}
\usepackage{tabularx}
\usepackage{array}
\usepackage{capt-of}
\usepackage{adjustbox}
\usepackage{amsmath}
\usepackage{xcolor}
\usepackage{etoolbox}
\usepackage{fontawesome}
\usepackage{amsfonts}
\usepackage{amssymb}

\usepackage{xcolor}
\definecolor{midnightblue}{rgb}{0.1, 0.1, 0.44}
\definecolor{firebrick}{rgb}{0.7, 0.13, 0.13}
\definecolor{darkcyan}{rgb}{0.0, 0.55, 0.55}

\definecolor{f1color}{RGB}{31,78,121}     
\definecolor{eccolor}{RGB}{46,125,50}     
\definecolor{fpcolor}{RGB}{156,87,0}      

\newcommand{\fones}[1]{\textcolor{f1color}{#1}}
\newcommand{\ecs}[1]{\textcolor{eccolor}{#1}}
\newcommand{\fps}[1]{\textcolor{fpcolor}{#1}}

\usepackage[dvipsnames]{xcolor}
\usepackage{pifont}
\newcommand{\cmark}{\textcolor{ForestGreen}{\ding{51}}} 
\newcommand{\xmark}{\textcolor{BrickRed}{\ding{55}}}   
\definecolor{DeepBlue}{RGB}{0,76,153}
\definecolor{SoftGray}{gray}{0.45}
\newcommand{\Fine}[1]{\textbf{\textcolor{NavyBlue}{#1}}}
\newcommand{\Over}[1]{\textbf{\textcolor{BurntOrange}{#1}}}

\newcommand{\auditcrit}[1]{\textbf{\textcolor{TealBlue}{#1}}}
\newcommand{\auditwarn}[1]{\textbf{\textcolor{Plum}{#1}}}

\definecolor{GoodGreen}{RGB}{0,120,80}
\definecolor{BadRed}{RGB}{160,40,40}

\newcommand{\good}[1]{\textcolor{GoodGreen}{\textbf{#1}}}
\newcommand{\bad}[1]{\textcolor{BadRed}{\textbf{#1}}}

\usepackage{enumitem}
\usepackage[most]{tcolorbox}
\title{PaperAudit-Bench: Benchmarking Error Detection in Research Papers for Critical Automated Peer Review\\[0.65em]
{\footnotesize\textcolor{orange}{\bfseries \faWarning\, WARNING: Contains intentionally AI-corrupted content for research use only.}}}

\author{
  Songjun Tu$^{1,2}$\thanks{Equal contribution.} \quad
  Yiwen Ma$^{1,2}$\footnotemark[1] \quad
  Jiahao Lin$^{1,2}$ \quad
  Qichao Zhang$^{1,2}$ \\
  \textbf{Xiangyuan Lan}$^{4}$ \quad
  \textbf{Junfeng Li}$^{1,3}$ \quad
  \textbf{Nan Xu}$^{3}$ \quad
  \textbf{Linjing Li}$^{1,2}$ \quad
  \textbf{Dongbin Zhao}$^{1,2}$\thanks{Corresponding author.}\\
  $^{1}$ Institute of Automation, Chinese Academy of Sciences \\
  $^{2}$ University of Chinese Academy of Sciences \\
  $^{3}$ Wenge Technology \quad
  $^{4}$ Pengcheng Laboratory\\
  \texttt{\{tusongjun2023,dongbin.zhao\}@ia.ac.cn}
}

\makeatletter

\newcommand{\Rmnum}[1]{\expandafter\@slowromancap\romannumeral #1@}
\makeatother

\begin{document}
\maketitle

\begin{abstract}
Large language models (LLMs) can generate fluent peer reviews but often miss sparse, subtle, and document-level scientific errors.
We introduce \textbf{PaperAudit-Bench}, a benchmark for evaluating paper-level error detection and error-aware automated review.
It comprises \textbf{(1) PaperAudit-Dataset}, 220 high-quality AI/ML papers containing more than 2,200 controlled synthetic errors across eight categories, and \textbf{(2) PaperAudit-Review}, an evidence-aware framework that links structured error detection to review generation.
Experiments across models, error categories, paper sections, and analysis depths show that sparse long-context auditing remains challenging for current LLMs.
Conditioning review generation on explicit, manuscript-grounded error evidence produces more conservative and discriminative assessments that align more closely with human judgments than representative automated-review baselines.
The dataset also supports post-training compact detectors through SFT and RL; Qwen3-14B achieves 55.8\% F1 at Detection@1 under a fixed held-out protocol.
A PubPeer transfer study and deployment in \textit{ScienceOne}, a scientific foundation-model platform led by the Chinese Academy of Sciences, further illustrate practical applicability.

\textbf{Code and Dataset Page:} \url{https://github.com/TU2021/PaperAudit}. 

\textbf{ScienceOne Manuscript Review:} \url{https://www.scienceone.cn/lit/\#/assistant}.
\end{abstract}

\begin{table*}[t]
  \centering
  \setlength{\tabcolsep}{3pt}
  \renewcommand{\arraystretch}{1.02}
  \begin{adjustbox}{width=\textwidth}
  \begin{tabular}{lllcccc}
  \toprule
  \textbf{Benchmark / Dataset} &
  \textbf{Source} &
  \textbf{Data Scale} &
  \textbf{Error Types} &
  \textbf{Detect Mode} &
  \textbf{Review Integration} &
  \textbf{Detector Training} \\
  \midrule
  Aspect-Perturb~\cite{li2025llms} &
  Synthetic &
  508 papers (79k edits) &
  \textbf{\textcolor{fpcolor}{S}} &
  \textbf{\textcolor{eccolor}{-}} &
  \cmark &
  \xmark \\
  
  SPOT~\cite{son2025ai} &
  Real &
  83 papers (91 errs) &
  \textbf{\textcolor{f1color}{M}} &
  \textbf{\textcolor{fpcolor}{S}} &
  \xmark &
  \xmark \\
  
  FLAWS~\cite{xi2025flaws} &
  Synthetic &
  713 papers (1 err / paper) &
  \textbf{\textcolor{fpcolor}{S}} &
  \textbf{\textcolor{fpcolor}{S}} &
  \xmark &
  \xmark \\
  
  PRISMM-Bench~\cite{selch2025prismm} &
  Real &
  353 papers (384 errs) &
  \textbf{\textcolor{fpcolor}{S}} &
  \textbf{\textcolor{fpcolor}{S}} &
  \xmark &
  \xmark \\
  
  DeepReview~\cite{zhu2025deepreview} &
  Real &
  13k papers &
  \textbf{\textcolor{eccolor}{-}} &
  \textbf{\textcolor{eccolor}{-}} &
  \cmark &
  \xmark \\
  
  \midrule
  \textbf{PaperAudit-Bench (Ours)} &
  Synthetic &
  220 papers ($>$10 errs / paper) &
  \textbf{\textcolor{f1color}{M}} &
  \textbf{\textcolor{f1color}{M}} &
  \cmark &
  \cmark \\
  \bottomrule
  \end{tabular}
  \end{adjustbox}
  \caption{
  Comparison of PaperAudit-Bench with existing benchmarks and review datasets.
  \textbf{\textcolor{f1color}{M}} denotes multi-type or multi-stage support,
  \textbf{\textcolor{fpcolor}{S}} denotes single-type or single-stage support,
  and \textbf{\textcolor{eccolor}{-}} denotes not applicable.
  }
  \label{tab:benchmark_comparison}
  \end{table*}

\section{Introduction}

Large language models (LLMs) are increasingly used for scientific writing and reviewing, making automated peer review a growing AI-for-Science topic~\cite{lu2024ai,chen2025ai4research,lin2023automated}.
Recent post-training and agentic workflows improve LLM reasoning and alignment, enabling adaptive-depth analysis and structured review generation~\cite{fu2025srft,tu2025learning,zhu2025deepreview,huang2026wese}.
Yet evidence indicates that AI-generated reviews often lack critical rigor, missing methodological flaws, cross-section inconsistencies, or overstated claims that human reviewers would flag~\cite{wu2024confidence,dycke2025automatic,li2025llms,zhu2025your}.

Despite these observations, such limitations remain difficult to assess because systematic benchmarks for \textit{paper-level} errors are scarce.
Such errors may appear locally within sections or as inconsistencies evident only when comparing distant parts of a paper, while large-scale collections of real papers with author-acknowledged mistakes remain rare.
Although peer-review resources such as NLPeer support computational analysis of review text~\cite{dycke2023nlpeer}, they lack controlled document-level error targets for evaluating whether AI reviewers can identify substantive flaws.
This gap motivates a benchmark for both diagnostic assessment of LLM error-detection capability and systematic study of how detection affects downstream review quality.

To address this gap, we introduce \textbf{PaperAudit-Bench}, a benchmark for auditing scientific papers and evaluating the critical assessment behavior of automated peer review systems.
PaperAudit-Bench consists of two components:
\textbf{\Rmnum{1}. PaperAudit-Dataset}, a \textbf{\textit{paper-level error corpus}} constructed by injecting diverse, sparse, and distributed errors into high-quality conference papers through model-driven synthetic editing, covering 8 major risk dimensions;
\textbf{\Rmnum{2}. PaperAudit-Review}, a unified framework with structured \textbf{\textit{error detection}} under three modes (fast single-pass, standard with document-level \textit{global memory}, and deeper multi-agent analysis) and \textbf{\textit{evidence-aware review}} for critical refinement.

Extensive experiments demonstrate that
(1) state-of-the-art LLMs exhibit substantial variability in detecting errors, with uneven sensitivity across error types and paper sections;
(2) structured detection and document-level global memory improve distributed-error coverage;
(3) PaperAudit-Dataset supports SFT and RL training of lightweight detectors, with Qwen3-14B reaching 55.8\% F1 at Detection@1;
(4) explicit error detection shifts AI reviewers toward more critical and discriminative evaluations aligned with human judgments;
(5) a zero-shot transfer study provides evidence on recent PubPeer-discussed papers; and
(6) deployment on the \textit{ScienceOne} platform demonstrates practical integration.
Overall, we aim to support, rather than replace, human reviewers with evidence-grounded automated auditing.

The main contributions are as follows:
\begin{tcolorbox}[
    enhanced,
    colback=black!2,
    colframe=black!15,
    boxrule=0.35pt,
    arc=1.5mm,
    sharp corners=southwest,
    sharp corners=southeast,
    left=1.2mm,right=1.2mm,top=1.2mm,bottom=1.2mm
]
\begin{enumerate}[leftmargin=1.5em, itemsep=2pt, topsep=2pt]
    \item We introduce \textbf{PaperAudit-Dataset}, a \textbf{paper-level error corpus} with over 2,200 annotated errors across 220 AI papers for sparse, multi-type, long-context error detection.
    \item We develop \textbf{PaperAudit-Review}, an \textbf{evidence-aware review framework} coupling structured error detection with critical review refinement.
    \item We provide broad empirical evidence across \textbf{model families}, \textbf{detection depths}, \textbf{human-alignment evaluation}, and \textbf{lightweight detector training}, demonstrating the difficulty and utility of paper-level auditing.
    \item We add a \textbf{real-error transfer study} and report deployment as the Manuscript Review module of \textit{ScienceOne}, showing practical integration for \textbf{pre-submission self-auditing}.
\end{enumerate}
\end{tcolorbox}

\section{Related Work}

\paragraph{AI-assisted Scientific Research.}
LLMs increasingly support the research lifecycle, including idea generation, experimentation, paper writing, and scientific agents~\cite{ren2025towards,chen2025ai4research,schmidgall2025agent,wang2025biologyllm,yamada2025ai,shao2025omniscientist,weng2025deepscientist,huang2026wese,xi2025llmagentssurvey}; reliable scientific reasoning and verification nevertheless remain challenging~\cite{chen2025mlr,kon2025exp,zhang2025mlrc,ren2025assisting,hou2025paperdebugger}.

\paragraph{AI-based Paper Review and Error Detection.}
Automated peer review has adopted LLMs for review generation, SFT/RL-based reviewing, and grounded feedback~\cite{yuan2022can,weng2025cycleresearcher,zhu2025deepreview,taechoyotin2025remor,zeng2025reviewrl}, but current systems still miss substantive errors and remain vulnerable to perturbations and prompt-level risks~\cite{wu2024confidence,zhou2024llm,li2025unveiling,li2025llms,zhu2025your,dycke2025automatic}.
Prior resources address real cases~\cite{son2025ai,zhang2025reviewing}, multimodal inconsistencies~\cite{selch2025prismm}, localized synthetic flaws~\cite{xi2025flaws}, robustness perturbations~\cite{li2025llms}, or AI-feedback settings~\cite{thakkar2025can,zhang2025reviewguard}, but do not jointly provide controlled, distributed, multi-type paper-level errors with integrated reviewing.
Table~\ref{tab:benchmark_comparison} situates \textbf{PaperAudit-Bench} along these dimensions; the full discussion is in Appendix~\ref{sec:appendix_related_work}.

\begin{figure*}[t]
  \includegraphics[width=\linewidth]{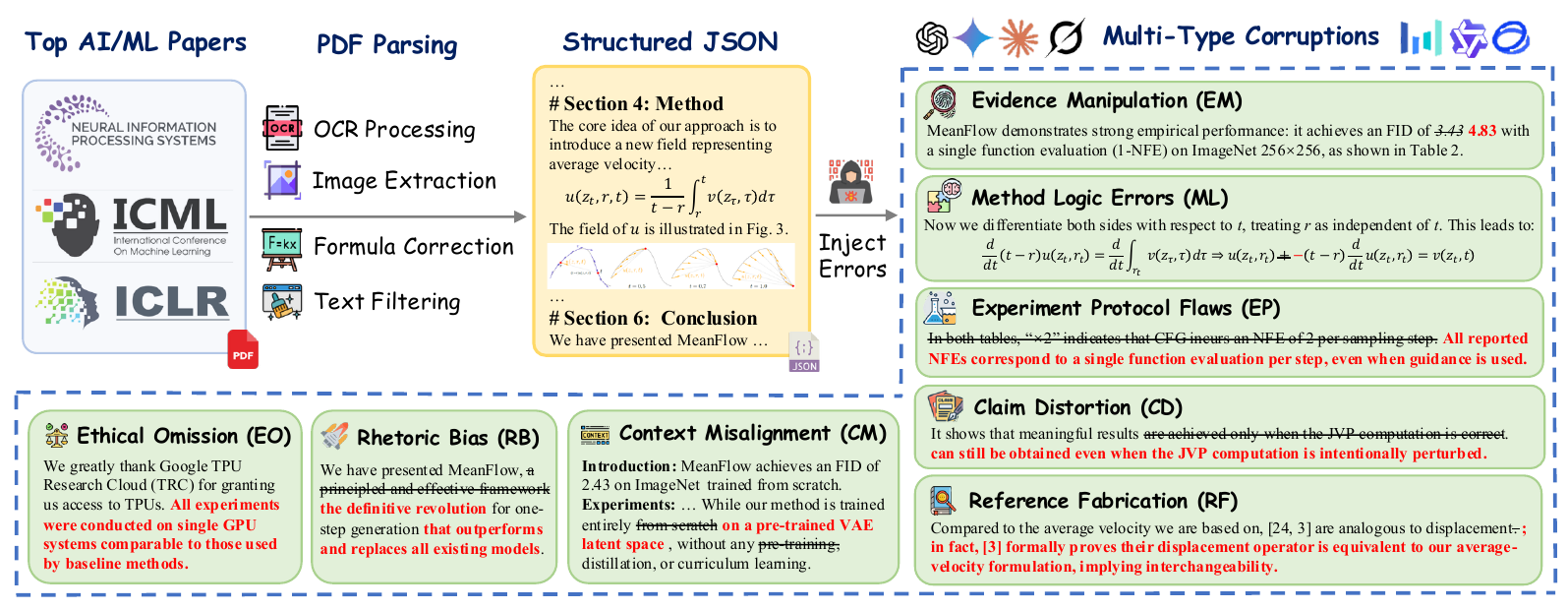}
  \caption{Overview of \textit{\textbf{PaperAudit-Dataset}} and examples of errors injected through \textit{model-driven synthetic editing}.}
  \label{fig:dataset}
\end{figure*}

\begin{figure*}[t]
  \includegraphics[width=\linewidth]{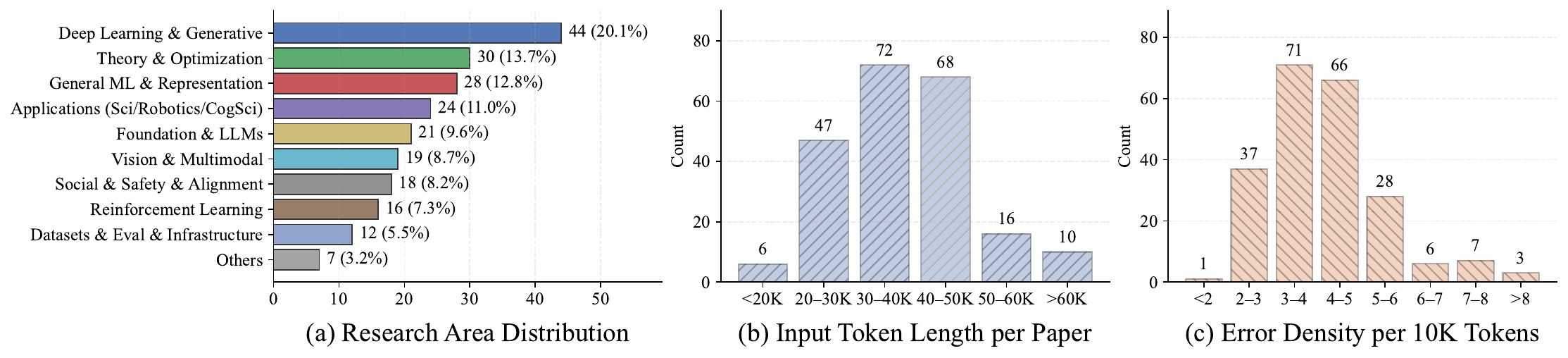}
    \caption{
    Statistics of \textit{\textbf{PaperAudit-Dataset}}. 
    (\textbf{a}) Distribution of research areas covered in PaperAudit-Dataset. 
    (\textbf{b}) Distribution of input token lengths per paper, estimated under the GPT-5 default configuration. 
    (\textbf{c}) Distribution of injected error density, measured as errors per 10K tokens, computed from GPT-5-generated corrupted papers.
    }
  \label{fig:dataset_stats}
\end{figure*}

\begin{figure*}[t]
  \centering
  \begin{subfigure}{\linewidth}
    \centering
    \includegraphics[width=\linewidth]{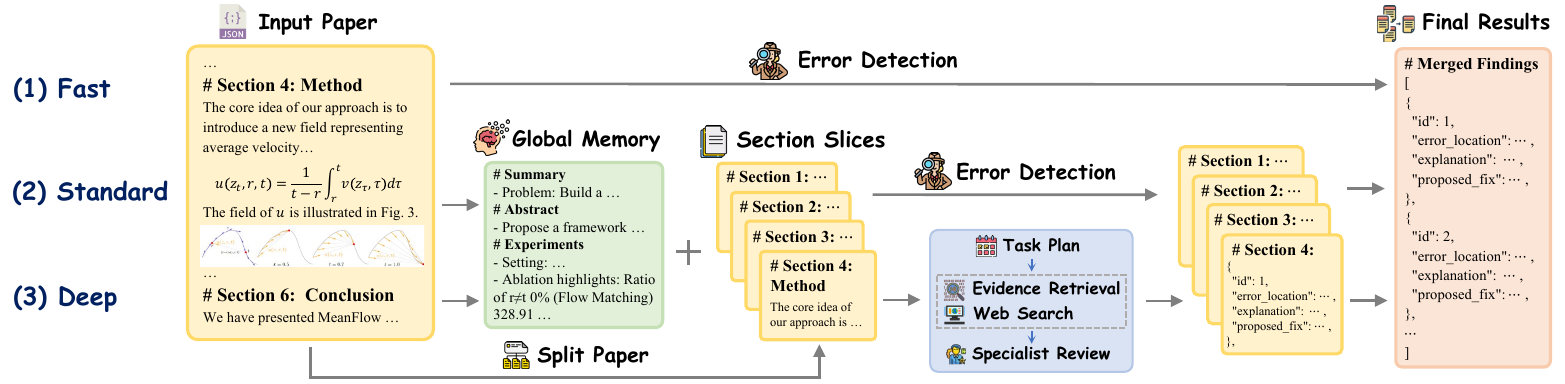}
    \caption{Detection workflow}
    \label{fig:detect-workflow}
  \end{subfigure}
  \begin{subfigure}{\linewidth}
    \centering
    \includegraphics[width=\linewidth]{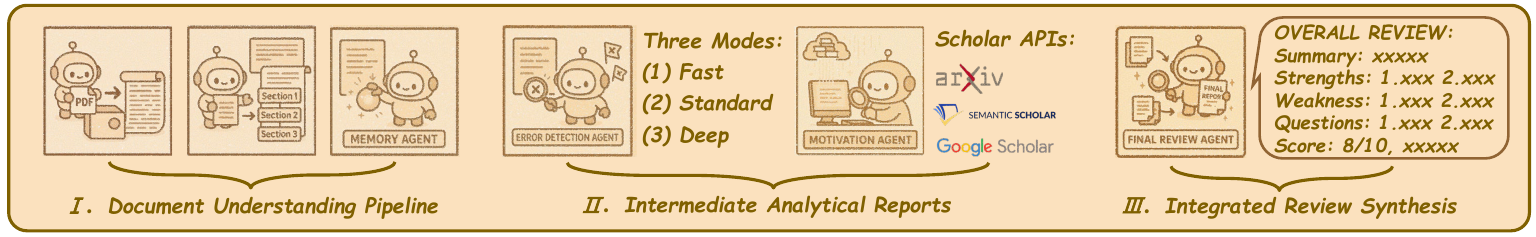}
    \caption{Review workflow}
    \label{fig:review-workflow}
  \end{subfigure}
  \caption{Overview of the \textbf{PaperAudit-Review Framework}. The detection workflow supports three levels of analytical depth, while the review workflow builds upon the detection results to synthesize critical assessments.}
  \label{fig:two-workflows}
\end{figure*}

\section{PaperAudit-Dataset: Benchmarking Error Detection for Research Papers}


\subsection{Paper Collection and Preprocessing}
\label{sec:dataset_source}

\textbf{PaperAudit-Dataset} is constructed from research papers accepted at the three top AI conferences:
\textbf{\textit{ICLR}}, \textbf{\textit{ICML}}, and \textbf{\textit{NeurIPS}} in \textbf{\textit{2025}}.
We focus on \textbf{\textit{oral}} papers, which are generally regarded as high-quality and contain relatively few obvious errors, ensuring that injected corruptions constitute the primary source of errors in the dataset.
Papers are collected from \textbf{\textit{OpenReview}} and filtered using a lightweight length-based criterion to exclude extremely long documents, which may introduce instability in long-context processing.
After filtering, the dataset comprises 107 ICLR, 67 ICML, and 46 NeurIPS papers in PDF format.

Each paper is converted into a \emph{structured, section-aware JSON representation} designed for multimodal language models.
The representation consists of an ordered sequence of text and figure blocks, each annotated with a semantic section label, providing a standardized input format for downstream error injection, detection, and review.
Implementation-level details of PDF parsing, multimodal extraction, section labeling, and rule-based cleanup are provided in Appendix~\ref{sec:appendix_data_construction}.

\subsection{Synthetic Error Injection}
\label{sec:dataset_injection}

To enable systematic evaluation of paper auditing capabilities, we adopt a \textit{model-driven synthetic editing} strategy to inject controllable paper-level errors.
For each source paper, multiple LLMs generate corrupted versions, with each model prompted to inject 10--20 errors across locations and sections. This is an \emph{evaluation-density design}: it provides enough labeled cases for stable category- and section-level analysis and should not be interpreted as an estimate of the natural error frequency in accepted papers.

Inspired by established guidelines on research integrity~\cite{national2017fostering,zhang2025reviewing}, we identify and design eight categories of errors common in papers, spanning from unintentional mistakes to more subtle forms of misrepresentation and misconduct. 
Detailed category descriptions are provided in Table~\ref{tab:appendix_error_types}.
Each patch is required to cover at least four sections to prevent error clustering, with all text spans non-overlapping to ensure statistical independence.
Patches that fail to match any verbatim original span or produce implausible or incoherent edits are filtered out via an automated post-hoc quality check; full filtering criteria and rejection statistics are detailed in Appendix~\ref{sec:appendix_error_injection}.
Importantly, many errors require paper-level reasoning across \textbf{distant sections} of a paper, rather than inspection of isolated text segments, such as verifying claims in the abstract against evidence in the experiments.
We find no significant variance in detection rates across different error loads or spatial distributions within a paper, indicating that the injected errors are effectively decoupled.

\subsection{Dataset Characteristics}
\label{sec:dataset_characteristics}

\textbf{PaperAudit-Dataset} is designed as a controlled stress test for sparse, paper-level, logically verifiable errors.
Figure~\ref{fig:dataset} summarizes the construction pipeline from paper collection and preprocessing to error injection, with representative examples for each error category.

\paragraph{Scale and Coverage.}
The dataset comprises \textbf{220 papers}: 107 from ICLR, 67 from ICML, and 46 from NeurIPS.
It covers a broad range of AI subfields, including generative modeling, optimization, representation learning, robotics, multimodal learning, reinforcement learning, and AI safety.
Each paper contains \textbf{approximately 15 injected errors} on average, of which \textbf{over 10 valid errors} are retained after filtering and quality control.

\paragraph{Long-Context and Sparse Distribution.}
As shown in Figure~\ref{fig:dataset_stats}, most papers contain roughly 20K--50K input tokens, making paper-level auditing a \textbf{long-context analysis problem}.
The retained errors are distributed across multiple sections and include both locally verifiable flaws and cross-section inconsistencies that require comparing distant claims, methods, and results.
Meanwhile, errors remain \textbf{sparse} with most papers containing only a few errors per 10K tokens.

\paragraph{Controlled Diagnostic Structure.}
Each error is associated with a predefined risk category, a document location, and an evidence-grounded explanation, enabling category-, section-, and model-level analysis rather than only an aggregate detection score.
The benchmark controls error type and location, but does not assume that synthetic corruptions are indistinguishable from naturally occurring author mistakes.

\section{PaperAudit-Review: Error Detection and Critical Review Framework}


\textbf{PaperAudit-Review} comprises an error detection component with varying analytical depth and a review component that integrates detected errors to support evidence-aware review, as summarized in Figure~\ref{fig:two-workflows}.

\subsection{Detection Workflow}
\label{sec:detect_workflow}
We organize detection into three \textbf{deployment-oriented cost--depth tiers} (Figure~\ref{fig:detect-workflow}), ranging from low-cost global screening to section-wise and multi-agent analysis.

\paragraph{Fast Mode: Single-Pass Global Detection.}
The \textit{\textbf{Fast}} mode performs a single-pass scan over the entire paper, treating the document as a whole without explicit section decomposition. It prioritizes prominent and high-level issues, particularly cross-section support mismatches, and is designed for efficient large-scale screening with minimal computational overhead.

\paragraph{Standard Mode: Section-Aware Detection with Global Context.}
The \textit{\textbf{Standard}} mode introduces explicit section awareness and a shared global memory.
The paper is decomposed into semantic sections, each reviewed independently while being conditioned on a global memory of the full document.
Global memory is constructed by prompting the \textit{same LLM} as the detection agent to produce a structured document summary (covering problem, contributions, experimental setup, and key claims), thereby maintaining reasoning consistency between memory construction and downstream section-level analysis.
This use of compact, structured context is related to recent memory-augmented agent work that organizes retrieved information into grounded evidence traces~\cite{ma2026memchain}.
This design enables finer-grained detection within individual sections while supporting cross-section reasoning through shared global memory, making it effective for identifying inconsistencies that are distributed across distant parts of the paper.

\paragraph{Deep Mode: Plan-Driven Multi-Agent Detection.}
The \textit{\textbf{Deep}} mode further extends the Standard setting by incorporating explicit planning and specialized analysis. It decomposes the review process into targeted tasks spanning multiple sections, optionally retrieves additional supporting evidence, and assigns each task to domain-specific specialist agents. This mode targets high-stakes, in-depth auditing at higher computational cost. Recent work on adaptive search-depth control similarly highlights the accuracy--efficiency trade-off in multi-step evidence acquisition~\cite{sun2026autosearch}.

Formal definitions, algorithmic details, and prompt specifications for all three detection modes are provided in Appendix~\ref{sec:appendix_error_detection}.

\subsection{Training Lightweight Detectors}
\label{sec:lightweight-training}

We further train compact detectors with a two-stage SFT+RL pipeline, targeting the large LLM detector in Section~\ref{sec:detect_workflow} as the main computational bottleneck and testing whether PaperAudit-Dataset can provide useful supervision beyond evaluation.

\paragraph{SFT with Structured Data.}
We first perform SFT on structured instances $\mathcal{D}_{\text{train}}=\{(x,y)\}$, where $x$ is a corrupted paper section and $y$ specifies the error type, location, and explanation.
The model is optimized with token-level cross-entropy over the annotation sequence, learning both error reasoning and output-format control.

\paragraph{RL with Multi-Dimensional Rewards.}
We then apply GRPO~\cite{shao2024deepseekmath} to improve detection quality and reduce redundant outputs.
The reward combines judge-estimated semantic precision, output conciseness, and penalties for abnormal behaviors such as empty responses or invalid JSON:
\begin{equation}
  \begin{aligned}
  &r(x, a, y) =
  \; w_p \cdot \text{Precision}(a, y) \\
  &+ w_c \cdot \text{Conciseness}(a)
  + \sum_{k=1}^{K} \mathcal{P}_k(a, y),
  \end{aligned}
  \end{equation}
This stage encourages focused, parsable findings and yields compact detectors for practical use in the workflow.
Implementation details are provided in Appendix~\ref{sec:appendix_training_details}.

\begin{figure}[b]
  \includegraphics[width=\linewidth]{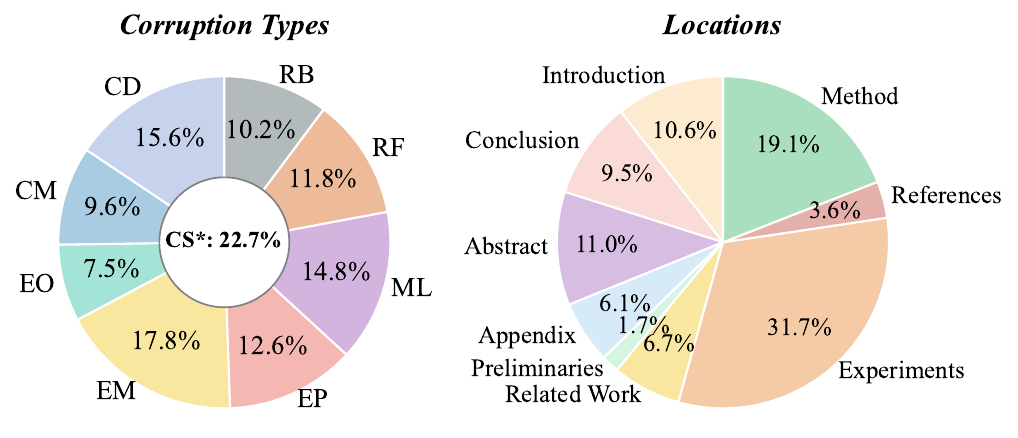}
  \caption{Distribution of injected error types and their locations in the \textit{NeurIPS} subset of \textit{PaperAudit-Dataset}. The statistics are averaged over corrupted papers generated by eight synthesis models.}
  \label{fig:nips25_corruption_location_pies}
\end{figure}

\begin{figure*}[t]
  \includegraphics[width=\linewidth]{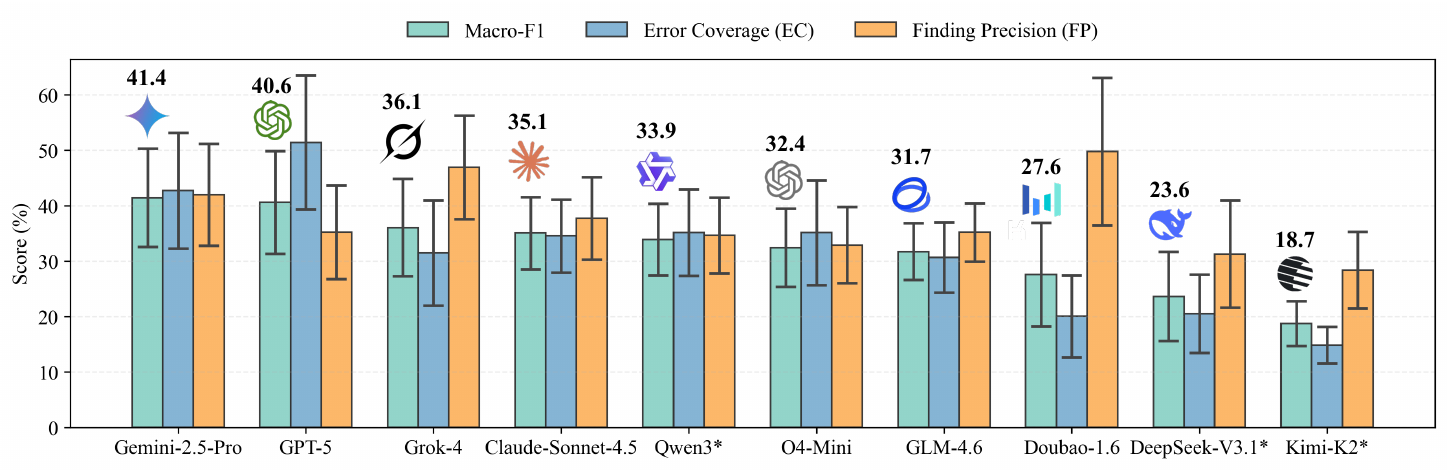}
  \caption{Fast-mode detection performance measured by Macro-F1, Error Coverage (EC), and Finding Precision (FP). Bars show mean and standard deviation across data synthesized by eight models; * denotes text-only inputs without visual content.}
  \label{fig:fast-mode-performance}
\end{figure*}

\subsection{Applying Error Detection to Peer Review}
\label{sec:review_pipeline}
Beyond providing actionable feedback to authors, error detection also enhances the \textbf{peer review} process by supporting more critical analysis.
Figure~\ref{fig:review-workflow} illustrates the overall review workflow, which integrates the proposed detection framework to enable \textbf{critical} and structured reviewing.
The process begins with document understanding, where the input paper is parsed into a structured representation and summarized into a global memory.
Based on this representation, a \textbf{baseline review} is first generated to provide an initial assessment.
In parallel, two analytical processes are conducted: \textbf{\textit{fine-grained error detection}} using the proposed detection workflow, and \textbf{\textit{novelty analysis}} that queries scholarly APIs to identify related work and assess similarity and originality.
The resulting error and motivation reports serve only as \textbf{silent attention cues}: the refinement stage cannot cite them as evidence, and every retained critique must be re-grounded in the manuscript with an explicit anchor. Minimal-edit and relevance-filtering constraints preserve the baseline review while suppressing stylistic, weakly supported, or low-impact critiques.
This full workflow underlies the Manuscript Review module deployed on \textit{ScienceOne}; the interface is shown in Appendix~\ref{sec:appendix_deployment}.

\section{Evaluation}
We evaluate \textit{PaperAudit} along three progressively connected research questions:

\begin{tcolorbox}[
    enhanced,
    colback=black!2,
    colframe=black!15,
    boxrule=0.35pt,
    arc=1.5mm,
    sharp corners=southwest,
    sharp corners=southeast,
    left=1.2mm,right=1.2mm,top=1.2mm,bottom=1.2mm
]
\begin{itemize}[leftmargin=8pt, itemsep=4pt, topsep=2pt, parsep=1pt]
  \item \textbf{Q1:} How detectable are paper-level errors across models, error types, and locations?
  \item \textbf{Q2:} Does explicit error detection lead to more critical AI-based peer reviews?
  \item \textbf{Q3:} Can PaperAudit-Dataset support training lightweight detectors?
\end{itemize}
\end{tcolorbox}

\subsection{Q1: Error Detectability across Models}

We first evaluate whether paper-level errors in \textit{PaperAudit-Dataset} are non-trivial and discriminative.

\paragraph{Dataset and Error Distribution.}
We evaluate on the \textit{NeurIPS} subset: 46 papers corrupted by eight synthesis models.
As shown in Figure~\ref{fig:nips25_corruption_location_pies}, errors cover diverse categories, appear in both technical and high-level sections, and include cross-section consistency errors (CS*) that require global context.
This subset provides enough variation across papers and synthesis models while keeping long-context evaluation tractable.
The CS* annotation is especially important because such errors often require comparing claims, methods, and evidence across distant sections rather than inspecting a local passage alone.

\begin{table*}[t]
  \centering
  \setlength{\tabcolsep}{4pt}
  \renewcommand{\arraystretch}{1.15}
  \begin{adjustbox}{max width=\textwidth,center}
  \begin{tabular}{lcccccc}
  \toprule
   & \textbf{GPT-5.1} & \textbf{Gemini-2.5-Pro} & \textbf{Grok-4} & \textbf{Claude-Sonnet-4.5} & \textbf{O4-Mini} & \textbf{Average} \\
  \midrule
  Human Detection Rate (\%) & 89.7 & 84.5 & 75.5 & 89.5 & 71.4 & \textbf{82.1} \\
  \bottomrule
  \end{tabular}
  \end{adjustbox}
  \caption{Human detection rate (\%) on the NeurIPS subset by synthesis model. PhD students judged whether each tampered passage was identifiable as an error from its surrounding source-paper context.}
  \label{tab:human_detection}
  \end{table*}

\begin{table*}[t]
  \centering
  \setlength{\tabcolsep}{4pt}
  \renewcommand{\arraystretch}{1.15}
  \begin{adjustbox}{width=\textwidth}
  \begin{tabular}{lccc}
  \toprule
  \textbf{Setting} & \textbf{Detected PubPeer Issues} & \textbf{Soft Issue Coverage} & \textbf{Paper Hit Rate} \\
  \midrule
  Fast (global scan) & 13 full + 11 partial & 24.34 & 36.00 \\
  Standard w/o global scan (memory-guided section audit) & 10 full + 13 partial & 21.71 & 34.00 \\
  Standard (global scan + memory-guided section audit) & \textbf{19 full + 14 partial} & \textbf{34.21} & \textbf{46.00} \\
  \bottomrule
  \end{tabular}
  \end{adjustbox}
  \caption{Zero-shot transfer on 50 recent PubPeer-discussed papers with 76 silver issue labels. Soft issue coverage and paper hit rate are percentages; coverage counts a full match as 1 and a partial match as 0.5.}
  \label{tab:pubpeer_transfer}
\end{table*}

\paragraph{Evaluation Metrics.}

We evaluate error detection using three complementary paper-level metrics capturing coverage, precision, and their trade-off.
Error Coverage (EC) measures coverage of injected ground-truth (GT) errors, while Finding Precision (FP) measures how many reported findings correspond to injected errors, defined as:

{\small
\[
\mathrm{EC} = \frac{\#\text{Matched Errors}}{\#\text{GT Errors}}, \ \
\mathrm{FP} = \frac{\#\text{Matched Findings}}{\#\text{All Findings}},
\]
}

Here, matched errors and findings are identified via \textbf{\textit{LLM-as-a-judge}} with GPT-5.1, which adjudicates semantic equivalence between detected findings and GT errors.
A judge-swap analysis using Gemini-2.5-Pro and Claude-Sonnet-4.5 changes absolute scores but preserves the relative detector ranking, supporting rank stability across judges (Appendix~\ref{sec:appendix_judge_swap}).
For each paper, we compute an F1 score as the harmonic mean of EC and FP, and report Macro-F1 by averaging across papers:
\[
\mathrm{Macro\text{-}F1} = \frac{1}{N} \sum_{i=1}^{N}
\frac{2 \cdot \mathrm{EC}_i \cdot \mathrm{FP}_i}{\mathrm{EC}_i + \mathrm{FP}_i}.
\]

Unlike standard classification settings, unmatched findings are not necessarily false positives, as even high-quality papers may surface reasonable but non-injected issues whose validity is difficult to conclusively determine.
Accordingly, some analyses place greater emphasis on Error Coverage (EC), which reflects the detector's ability to identify injected errors.
Potential over-coverage is primarily examined through qualitative \textit{case study} analysis.

\paragraph{Detection Models Exhibit Distinct Coverage-Precision Trade-offs.}

Figure~\ref{fig:fast-mode-performance} summarizes \textit{Fast}-mode detection.
Gemini-2.5-Pro and GPT-5 obtain the highest Macro-F1 scores (41.4\% and 40.6\%); GPT-5 reaches the highest EC (51.4\%) by reporting more findings, while Gemini-2.5-Pro is more balanced in FP.
More conservative models such as Doubao-Seed-1.6 and Kimi-K2 produce fewer findings and lower coverage.
This reveals a clear coverage--precision trade-off: stronger models are more willing to flag subtle issues, whereas conservative models miss many sparse errors under long-context inputs.
Detailed per-model statistics are in Appendix~\ref{sec:appendix_fast_mode}.

\paragraph{Cost of the Full Fast-Mode Grid.}
The full evaluation grid contains $10\times8\times46=3{,}680$ detector runs and 3{,}680 GPT-5.1 adjudication calls. Based on the observed NeurIPS-subset traces, this corresponds to approximately 100--135M input tokens and 12--20M output tokens. At public API prices, excluding provider-specific discounts, the estimated total inference and adjudication cost is \textbf{USD 250--400}; the range reflects provider pricing, image-token accounting, and retry rates.

\begin{figure}[t]
  \includegraphics[width=\linewidth]{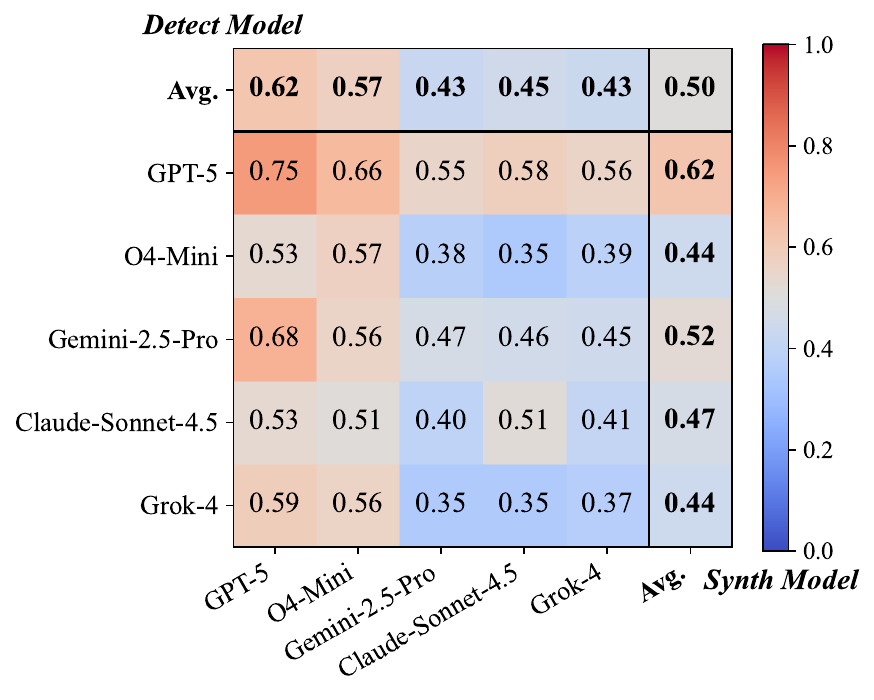}
  \caption{Error Coverage under the Standard mode across synthesis and detection model combinations. Rows correspond to detection models and columns to synthesis models. Row-wise and column-wise averages summarize overall detection capability and error detectability, respectively.}
  \label{fig:hotmap}
\end{figure}

\paragraph{Error Detectability Varies across Synthesis Models and Reflects Detector-Specific Biases.}

Figure~\ref{fig:hotmap} shows that detectability also depends on the synthesis model: GPT-5 errors are most detectable on average, while Grok-4 errors are more concealed.
We also observe mild same-model detection bias, likely from stylistic artifacts.
Therefore, all primary results average over eight synthesis models to reduce generator-specific effects.
This aggregation is important because otherwise the benchmark could overestimate reasoning ability by rewarding model-specific editing traces rather than robust scientific auditing.

\paragraph{Detectability Depends on Semantics and Location.}

\begin{figure}[t]
  \centering
  \includegraphics[width=\linewidth]{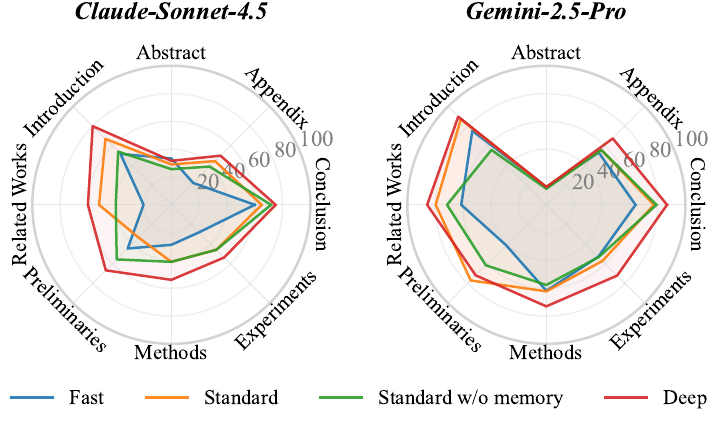}
  \caption{Error coverage across paper sections under different detection settings. Radial axes report EC (\%); ``Standard w/o memory'' retains section-wise auditing but removes the shared global memory.}
  \label{fig:section_coverage}
\end{figure}

Figure~\ref{fig:section_coverage} shows that error coverage improves with detection depth but varies by location; the type-level analysis further shows that rhetorical errors are harder to surface, whereas explicit inconsistencies are easier to identify.
Removing \textit{global memory} particularly hurts cross-section errors, and abstract errors remain difficult because high-level claims require later evidence.
Full type- and location-level results are in Appendix~\ref{sec:appendix_detectability} and \ref{sec:appendix_case_study_detect}.

\paragraph{Human Study Validates Logical Identifiability.}
Five PhD students with ML/NLP research backgrounds judged whether each tampered passage was identifiable as an error from its original paper context, without external knowledge.
As shown in Table~\ref{tab:human_detection}, they identify \textbf{82.1\%} of injected errors on average, supporting the logical groundedness and identifiability of the benchmark rather than indistinguishability from natural author errors.
Full task and participant details are provided in Appendix~\ref{sec:human_verification_protocol}.

\paragraph{Transfer to PubPeer-discussed Papers.}
To complement the controlled synthetic benchmark, we conduct a zero-shot transfer study on 50 recent PubPeer-discussed papers with accessible PDFs and 76 PubPeer-derived silver issue labels, using GPT-5-2025-08-07 as the detector.
Table~\ref{tab:pubpeer_transfer} reports exact/partial issue matches and paper-level hits. The combined Standard setting achieves \textbf{34.21\% soft issue coverage} and identifies at least one issue in \textbf{46.00\% of papers}. Among the 42 labels checkable from manuscript text alone, it reaches \textbf{42.86\% soft coverage} (14 full and 8 partial matches), providing a real-error reference point beyond synthetic edits.

\subsection{Q2: Error-Aware AI Peer Review}
\label{sec:alignment_human}

We next test whether explicit error detection improves AI peer review.
Following Section~\ref{sec:review_pipeline}, we refine a standard AI-reviewing baseline\footnote{\url{https://github.com/NeuroDong/Ai-Review}} with error and motivation reports to produce \textit{PaperAudit-Review}; \textit{DeepReview}~\cite{zhu2025deepreview} serves as a multi-perspective comparison.

\begin{figure*}[t]
  \centering
  \includegraphics[width=\textwidth]{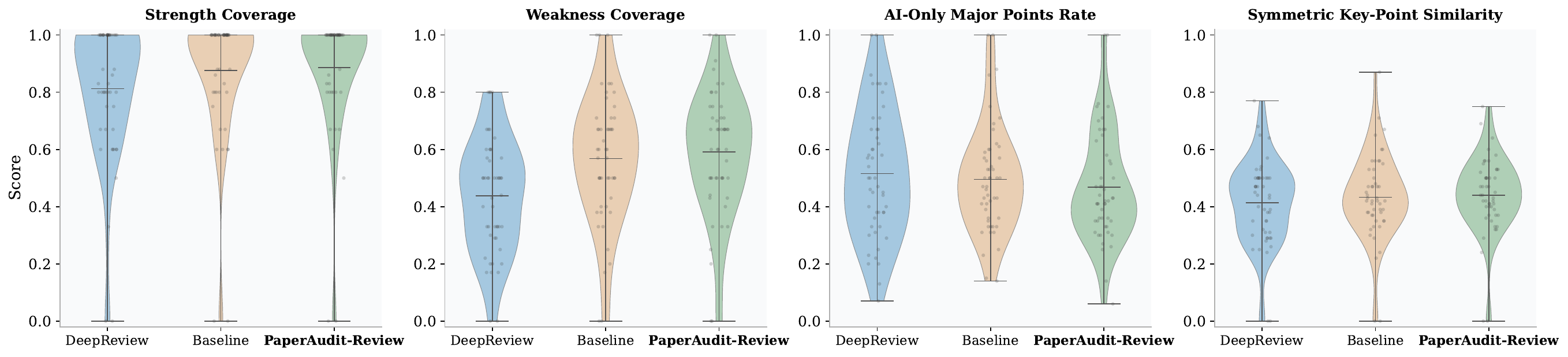}
  \caption{Per-paper alignment distributions. \textit{PaperAudit-Review} better covers human-identified weaknesses while controlling AI-only extra points.}
  \label{fig:method_coverage_comparison}
  \end{figure*}

\paragraph{PaperAudit-Review Produces More Critical and Discriminative Peer Reviews.}
Using 10-point review scores, corrupted papers receive lower ratings than originals, especially in technical quality and clarity (Appendix~\ref{sec:appendix_q2_additional}; Table~\ref{tab:vertical_review_comparison}).
Compared with the baseline, PaperAudit-Review assigns stricter diagnostic scores because explicit error evidence makes critiques more concrete~\cite{jin2024agentreview}.
This stricter behavior also affects original papers, where the detector may surface minor but substantively relevant weaknesses rather than only reacting to injected errors.
Its main value is more demanding, evidence-grounded feedback, supported by qualitative cases and lexical analysis in Appendix~\ref{sec:appendix_case_study_review} and Appendix~\ref{sec:appendix_q2_additional}.
Score-drop analysis further shows that method-logic contradictions and distorted claims drive most reductions, while presentation- or protocol-level issues have weaker effects (Appendix~\ref{sec:appendix_review_regression}).

\paragraph{PaperAudit-Review Exhibits Improved Alignment with Human Judgments.}

We further evaluate alignment on 50 randomly sampled ICLR~2026 submissions using score-based ranking metrics and coverage-based review-text metrics.
The former measures relative ordering consistency with averaged human scores, while the latter asks whether AI reviews cover the substantive strengths and weaknesses emphasized by human reviewers.
As shown in Table~\ref{tab:human_alignment_combined}, \textit{PaperAudit} improves over the baseline and \textit{DeepReview} across both views, with lower rank error, higher agreement, stronger weakness coverage, fewer AI-only major points, and higher symmetric coverage.
Although absolute alignment remains limited, the gains indicate a structural shift toward evidence-anchored critique rather than merely more verbose reviewing.
The adjudication step further helps filter low-specificity nitpicks, so the improvement reflects more targeted criticism rather than indiscriminate issue generation.
Figure~\ref{fig:method_coverage_comparison} shows this distributionally: \textit{PaperAudit} more consistently covers human-identified weaknesses while controlling extra AI-only points.
Metric details are in Appendix~\ref{sec:human_alignment_metrics} and \ref{sec:coverage_alignment_prompt}.

\begin{table}[t]
\centering
\setlength{\tabcolsep}{2pt}
\renewcommand{\arraystretch}{1.15}
\begin{adjustbox}{width=\columnwidth}
\begin{tabular}{lcccc}
\toprule
\multicolumn{5}{l}{\textbf{Panel A: Score-based Alignment with Human Reviewers}} \\
\midrule
\textbf{Method} &
\textbf{R-MSE} \textbf{\textcolor{eccolor}{$\downarrow$}} &
\textbf{Spearman} \textbf{\textcolor{fpcolor}{$\uparrow$}} &
\textbf{Kendall} \textbf{\textcolor{fpcolor}{$\uparrow$}} &
\textbf{P-Acc} \textbf{\textcolor{fpcolor}{$\uparrow$}} \\
\midrule
\textbf{\textit{Baseline}}   & 0.152 & 0.119 & 0.085 & 0.545 \\
\textbf{\textit{PaperAudit}} & \textbf{0.148} & \textbf{0.142} & \textbf{0.109} & \textbf{0.557} \\
\textbf{\textit{DeepReview}} & 0.160 & 0.075 & 0.057 & 0.529 \\
\midrule\midrule
\multicolumn{5}{l}{\textbf{Panel B: Coverage-based Alignment with Human Reviewers}} \\
\midrule
\textbf{Method} &
\textbf{Str-Cov} \textbf{\textcolor{fpcolor}{$\uparrow$}} &
\textbf{Weak-Cov} \textbf{\textcolor{fpcolor}{$\uparrow$}} &
\textbf{AI-Extra} \textbf{\textcolor{eccolor}{$\downarrow$}} &
\textbf{Sym-Cov} \textbf{\textcolor{fpcolor}{$\uparrow$}} \\
\midrule
\textbf{\textit{Baseline}}   & 0.876 & 0.568 & 0.495 & 0.433 \\
\textbf{\textit{PaperAudit}} & \textbf{0.886} & \textbf{0.591} & \textbf{0.468} & \textbf{0.444} \\
\textbf{\textit{DeepReview}} & 0.844 & 0.482 & 0.488 & 0.412 \\
\bottomrule
\end{tabular}
\end{adjustbox}
\caption{Alignment with human reviewers on 50 ICLR~2026 submissions. Panel A reports score-based alignment; Panel B reports judge coverage alignment.}
\label{tab:human_alignment_combined}
\end{table}

\subsection{Q3: Evaluating Lightweight Detectors}
\label{sec:rq3}

Finally, we test whether \textit{PaperAudit-Dataset} can train compact detectors.

\paragraph{Dataset and Settings.}
We train \textit{Llama3.2-3B}, \textit{Qwen3-8B}, and \textit{Qwen3-14B} on 5{,}161 ICLR/NeurIPS synthetic instances using SFT+RL, and evaluate on 1{,}209 ICML corrupted section-level samples mainly synthesized by \textit{GPT-5} and \textit{O4-Mini}.
The training set covers eight synthesis models to reduce overfitting to any single generator, while the ICML branch serves as a held-out conference split.
Detection@1 evaluates the highest-ranked finding per sample, whereas Detection@all evaluates the complete returned finding set.
Detailed hyperparameters and judge settings are provided in Appendix~\ref{sec:appendix_training_details}.

\begin{table}[t]
\centering
\renewcommand{\arraystretch}{1.15}
\setlength{\tabcolsep}{4pt}
\begin{adjustbox}{width=0.95\linewidth}
\begin{tabular}{lcccccc}
\toprule
 & \multicolumn{3}{c}{\textbf{Detection@1}}
 & \multicolumn{3}{c}{\textbf{Detection@all}} \\
\cmidrule(lr){2-4} \cmidrule(lr){5-7}
\textbf{Model} & EC & FP & F1 & EC & FP & F1 \\
\midrule
\textit{Gemini-2.5-Pro} & 48.0 & 53.8 & 50.6 & 75.4 & 27.3 & 40.1 \\
\textit{Claude-Sonnet-4.5} & 45.8 & 51.0 & 48.3 & \textbf{82.0} & 20.5 & 32.8 \\
\textit{Qwen3-235B-A22B} & 43.5 & 48.6 & 45.9 & 79.1 & 20.6 & 32.6 \\
\textit{DeepSeek-V3.1} & 29.0 & 32.3 & 30.6 & 67.1 & 18.6 & 29.1 \\
\midrule
\textbf{Llama3.2-3B} & 6.4 & 7.1 & 6.7 & 34.2 & 5.6 & 10.1 \\
\quad + SFT & 27.3 & 30.4 & 28.7 & 36.8 & 41.0 & 38.8 \\
\quad + SFT + RL & 30.4 & 33.9 & 32.1 & 37.8 & 42.1 & 39.8 \\
\midrule
\textbf{Qwen3-8B} & 13.9 & 20.5 & 16.6 & 40.5 & 38.5 & 39.5 \\
\quad + SFT & 38.4 & 42.8 & 40.5 & 43.1 & 47.6 & 45.2 \\
\quad + SFT + RL & 52.3 & 58.3 & 55.1 & 56.6 & 63.1 & 59.7 \\
\midrule
\textbf{Qwen3-14B} & 16.9 & 20.5 & 18.5 & 43.7 & 40.5 & 42.0 \\
\quad + SFT & 42.4 & 47.2 & 44.7 & 48.5 & 52.7 & 50.5 \\
\quad + SFT + RL & \textbf{52.9} & \textbf{59.0} & \textbf{55.8} & 57.1 & \textbf{63.7} & \textbf{60.2} \\
\bottomrule
\end{tabular}
\end{adjustbox}
\caption{Detection performance of general-purpose and post-trained lightweight detectors under Detection@1 and Detection@all.}
\label{tab:detection}
\end{table}

\paragraph{Performance.}
Table~\ref{tab:detection} shows that SFT and RL consistently improve all three backbones.
Qwen3-8B after SFT+RL becomes comparable to larger general-purpose models, and Qwen3-14B reaches the best Detection@1 F1 (55.8\%) under our fixed held-out protocol.
Under Detection@all, frontier models obtain high EC but low FP by over-reporting, whereas post-trained detectors achieve a more balanced EC--FP trade-off.
In a supplementary smaller-backbone stability check, Qwen3-1.7B SFT+RL yields $30.08\!\pm\!0.55$ EC, $33.56\!\pm\!0.65$ FP, and $31.70\!\pm\!0.62$ F1 across three seeds; reward-weight ablations are similarly stable (Appendix~\ref{sec:appendix_training_details}).

\subsection{Case Study and Deployment}
Appendix~\ref{sec:appendix_case_study} provides qualitative detection and review cases, plus the deployment of PaperAudit-Review as the Manuscript Review module of \textit{ScienceOne}, a scientific foundation-model platform led by the Chinese Academy of Sciences.

\section{Conclusion}
We present PaperAudit-Bench for paper-level error detection and error-aware automated review.
Experiments show that sparse, distributed errors remain challenging, while structured detection, global context, and post-trained compact detectors improve evidence-grounded auditing; transfer to PubPeer-discussed papers provides a complementary real-error reference point.
Deployment on \textit{ScienceOne} further demonstrates practical utility for assistive pre-submission self-auditing.

\clearpage

\section*{Acknowledgments}
This work is supported by the Beijing Nova Program 202604841268 and Strategic Priority Research Program of Chinese Academy of Sciences under Grant XDA0480302.

\section*{Limitations}
We note three directions for broader validation and extension.
\begin{itemize}[leftmargin=*, itemsep=0pt]

\item \textbf{Domain and language coverage.}
PaperAudit-Bench currently focuses on English AI/ML papers. Extending it to other disciplines, languages, and multimodal evidence would broaden its applicability and may require domain-specific error taxonomies and evaluation criteria.

\item \textbf{Controlled error construction.}
Synthetic injection provides reproducible labels and systematic coverage, but does not capture every form of naturally occurring scientific error. Our PubPeer transfer study offers a complementary real-error reference point, although these annotations are not exhaustive.

\item \textbf{Evaluation configurations.}
Our experiments use fixed prompts, model settings, and evaluation protocols to ensure comparability. Future work can expand the set of detector and judge models, include larger-scale human evaluation, and examine additional retrieval and calibration strategies.

\end{itemize}

\section*{Ethical Considerations}

This work uses LLMs to synthetically modify research papers and generate automated reviews solely for benchmark construction and evaluation. 
PaperAudit-Bench is a defensive resource for evaluating detectors and strengthening research integrity, not a tool for manuscript manipulation.
We \textbf{DO NOT} encourage academic misconduct, \textbf{NOR DO} we advocate replacing human reviewers in real peer-review processes.
Responsible release should preserve the corruption warning, separate corrupted documents from source papers, and document intended benchmark and self-auditing uses.
Document-level prompt injection, adversarial formatting, and parsing failures remain risks: structured JSON and evidence-anchor constraints mitigate but do not eliminate them.
LLM-based agents should be used only as assistive tools, with final judgments reserved for human experts.

\bibliography{custom}

\clearpage
\appendix
\section{Related Work Details}

\paragraph{AI-assisted Scientific Research}
\label{sec:appendix_related_work}
Recent advances have enabled AI-assisted scientific research across the research lifecycle, including idea formulation, experimentation, and paper writing \cite{ren2025towards, chen2025ai4research, schmidgall2025agent}.
This line of work includes end-to-end research automation systems such as \textit{The AI Scientist} \cite{lu2024ai} and its extensions \cite{yamada2025ai}, as well as specialized scientific agents for problem solving and researcher training \cite{weng2025deepscientist, shao2025omniscientist}.
Scientific idea generation has also been explored through chaining strategies \cite{li2024chain}, self-reflection-guided beam search \cite{hu2025nova}, and Monte Carlo tree search \cite{garikaparthi2025iris}, alongside benchmarks for evaluating idea diversity and quality \cite{ruan2024liveideabench, qiu2025ai}.
Despite this progress, autonomously conducting complex scientific experiments remains challenging for current systems \cite{chen2025mlr, kon2025exp, zhang2025mlrc}, whereas LLMs have demonstrated strong capability in drafting academic papers when provided with existing materials and results \cite{ren2025assisting, hou2025paperdebugger}.
Complementary to AI research systems, we focus on paper review, emphasizing critical assessment and error-aware feedback for refinement.

\paragraph{AI-based Paper Review}
Recent work on automated peer review has been framed within broader efforts to advance automated research workflows~\cite{weng2025cycleresearcher}, training LLMs via SFT and RL for human-like deep reasoning~\cite{zhu2025deepreview,taechoyotin2025remor} and for generating comprehensive, factually grounded reviews~\cite{zeng2025reviewrl}.
However, accumulating evidence shows that current AI reviewers have limited critical understanding, often missing substantive weaknesses in scientific papers~\cite{wu2024confidence,zhou2024llm,li2025unveiling}. They are also susceptible to perturbations in papers or rebuttals~\cite{li2025llms} and frequently fail to detect flawed reasoning, leading to biased or unreliable review judgments~\cite{dycke2025automatic}.
To address these limitations, some studies evaluate LLMs on papers with author-acknowledged errors from \textit{Withdrawn arXiv} and \textit{PubPeer}~\cite{son2025ai,zhang2025reviewing}, while parallel efforts improve review reliability via AI feedback~\cite{thakkar2025can} or synthetic-data-based detection of deficient peer reviews~\cite{zhang2025reviewguard}.
In contrast, we introduce a comprehensive dataset with accompanying error detection workflows that span multiple error types, supporting more critical peer review under perturbations.

\paragraph{Comparing to Existing Benchmarks}
Existing benchmarks for paper review and error analysis differ substantially along data provenance, error granularity, task formulation, and workflow design, as summarized in Table~\ref{tab:benchmark_comparison}.
SPOT~\cite{son2025ai} and PRISMM-Bench~\cite{selch2025prismm} are constructed from real review ecosystems, drawing author-confirmed errors from Withdrawn arXiv and PubPeer or reviewer-flagged multimodal inconsistencies from OpenReview.
While both operate at the document level, their supervision remains task-specific: SPOT evaluates error-existence verification over a small set of author-confirmed cases, and PRISMM formulates cross-modal inconsistencies as structured QA tasks, without modeling distributed or heterogeneous errors across a paper.
FLAWS~\cite{xi2025flaws} adopts a synthetic setting by injecting a single claim-invalidating flaw into each paper, producing paper--error pairs with explicit span-level localization supervision.
In contrast, Aspect-Guided Multi-Level Perturbation Analysis~\cite{li2025llms} is not an error annotation benchmark but a robustness analysis framework, applying controlled perturbations to papers, reviews, and rebuttals to study reviewer sensitivity, without providing error labels or detection targets.
Overall, prior resources either rely on scarce real error cases, focus on single-error or localized verification, or analyze review robustness without explicit error supervision.
PaperAudit-Bench complements these efforts by introducing controlled, document-level errors spanning multiple categories and distributed across full papers, together with multi-stage detection and explicit integration into automated review workflows.

\section{Additional Evaluation Details}

\subsection{Human Verification Study Protocol}
\label{sec:human_verification_protocol}

We conducted a lightweight expert verification study to assess whether the injected corruptions are logically identifiable as errors when their original context is provided.
The study involved five PhD students with relevant ML/NLP research background, recruited as expert annotators through academic contacts.
They were not recruited through a paid crowdsourcing platform, and no separate monetary compensation was provided for this short voluntary expert annotation task.

\paragraph{Instructions Given to Participants.}
Participants received the following instructions:
\begin{tcolorbox}[
  enhanced,
  breakable,
  colback=black!2,
  colframe=black!25,
  boxrule=0.4pt,
  arc=1.5mm,
  left=1.5mm,
  right=1.5mm,
  top=1.2mm,
  bottom=1.2mm
]
\small
You will be shown a tampered passage from an AI research paper together with its original context.
Your task is to judge whether the tampered passage can be identified as an error using only the provided paper context, without relying on external knowledge.
Please mark each case as identifiable or not identifiable, and focus on substantive inconsistencies such as contradicted claims, mismatched experimental descriptions, invalid methodological logic, or unsupported conclusions.
Do not attempt to infer author identity, venue decisions, or any personal information.
Your judgments will be used only in aggregate form to evaluate whether the injected errors are logically identifiable from the manuscript context.
This is a minimal-risk expert verification task over publicly available scientific papers and synthetic edits; no personal, sensitive, demographic, medical, or behavioral information will be collected.
\end{tcolorbox}

\paragraph{Consent, Privacy, and Ethics.}
Annotators were informed before participation that their judgments would be used only in aggregate form for benchmark evaluation.
No personal, sensitive, demographic, medical, or behavioral information was collected, and no individual annotator responses are reported.
No ethics review board approval was obtained.
The study involved no deception or intervention and consisted only of expert verification over publicly available scientific papers and synthetic edits.

\subsection{LLM-Judge Stability: Judge-Swap Experiment}
\label{sec:appendix_judge_swap}
To assess judge sensitivity, we conduct a judge-swap experiment in which the same set of detection results (Fast mode, NeurIPS subset) is re-evaluated using three different frontier judge models.
Table~\ref{tab:judge_swap} reports Macro-F1 (\%) across five detection models evaluated by each judge.
Although absolute scores vary across judges (Claude-Sonnet-4.5 is more permissive than GPT-5.1), the \textit{relative ranking of detectors remains consistent} across all judges, with Gemini-2.5-Pro and GPT-5 consistently outperforming others.
This experiment supports \emph{rank stability}, but does not establish absolute matching correctness.

\begin{table}[h]
\centering
\setlength{\tabcolsep}{1.5pt}
\renewcommand{\arraystretch}{1.05}
\begin{adjustbox}{max width=\columnwidth,center}
\begin{tabular}{lccccc}
\toprule
\textbf{Judge} & \textbf{Gemini-2.5} & \textbf{GPT-5} & \textbf{Grok-4} & \textbf{Claude-4.5} & \textbf{O4-Mini} \\
\midrule
GPT-5.1           & 41.4 & 40.6 & 36.1 & 35.1 & 32.4 \\
Gemini-2.5-Pro    & 44.8 & 41.2 & 39.0 & 38.9 & 35.4 \\
Claude-Sonnet-4.5 & 59.5 & 60.6 & 48.9 & 49.1 & 49.8 \\
\midrule
Average           & \textbf{48.6} & \textbf{47.5} & 41.3 & 41.0 & 39.2 \\
\bottomrule
\end{tabular}
\end{adjustbox}
\caption{Judge-swap experiment: Macro-F1 (\%) of five detection models (columns) evaluated by three judge models (rows) under Fast mode. The relative ranking of detectors is consistent across all judges.}
\label{tab:judge_swap}
\end{table}

\subsection{Fast-Mode Detection Performance Breakdown}
\label{sec:appendix_fast_mode}
Table~\ref{tab:fast_mode_split} provides a comprehensive breakdown of detection performance under the \textit{Fast} mode across detection models and synthesis models.
Each entry reports Macro-F1, Error Coverage (EC), and Finding Precision (FP), highlighting distinct detection behaviors across models.
Stronger models such as GPT-5 and Gemini-2.5-Pro achieve higher overall effectiveness, while lighter or more conservative models tend to trade coverage for precision by producing fewer findings.
This detailed analysis complements the aggregate trends discussed in the main text.

\begin{table*}[t]
\centering
\scriptsize
\setlength{\tabcolsep}{4pt}
\renewcommand{\arraystretch}{1.15}

\begin{adjustbox}{width=.85\textwidth}
\begin{tabular}{lcccc}
\toprule
\textbf{Detect $\backslash$ Synth}
& \textbf{GPT-5}
& \textbf{O4-Mini}
& \textbf{Gemini-2.5}
& \textbf{Claude-4.5} \\
\midrule
\textbf{Gemini-2.5-Pro} & \fones{60.5}\ / \ \ecs{57.9}\ / \ \fps{64.1} & \fones{44.6}\ / \ \ecs{51.3}\ / \ \fps{40.2} & \fones{39.7}\ / \ \ecs{38.5}\ / \ \fps{42.3} & \fones{35.4}\ / \ \ecs{35.5}\ / \ \fps{36.5} \\
\textbf{GPT-5}          & \fones{62.5}\ / \ \ecs{72.4}\ / \ \fps{56.0} & \fones{42.1}\ / \ \ecs{60.8}\ / \ \fps{33.6} & \fones{37.4}\ / \ \ecs{44.7}\ / \ \fps{33.5} & \fones{38.7}\ / \ \ecs{46.5}\ / \ \fps{34.5} \\
\textbf{Grok-4}         & \fones{48.1}\ / \ \ecs{39.4}\ / \ \fps{66.3} & \fones{45.2}\ / \ \ecs{43.4}\ / \ \fps{50.1} & \fones{31.6}\ / \ \ecs{25.8}\ / \ \fps{43.3} & \fones{35.1}\ / \ \ecs{29.4}\ / \ \fps{43.6} \\
\textbf{Claude-4.5}     & \fones{43.1}\ / \ \ecs{39.4}\ / \ \fps{49.0} & \fones{36.1}\ / \ \ecs{39.2}\ / \ \fps{34.6} & \fones{29.9}\ / \ \ecs{28.5}\ / \ \fps{33.1} & \fones{47.3}\ / \ \ecs{46.3}\ / \ \fps{50.2} \\
\textbf{Qwen3*}         & \fones{45.7}\ / \ \ecs{43.3}\ / \ \fps{49.6} & \fones{40.5}\ / \ \ecs{46.2}\ / \ \fps{36.9} & \fones{27.6}\ / \ \ecs{26.5}\ / \ \fps{29.9} & \fones{36.2}\ / \ \ecs{36.8}\ / \ \fps{38.4} \\
\textbf{O4-Mini}        & \fones{45.3}\ / \ \ecs{45.0}\ / \ \fps{48.9} & \fones{39.2}\ / \ \ecs{46.9}\ / \ \fps{35.5} & \fones{27.8}\ / \ \ecs{27.8}\ / \ \fps{29.8} & \fones{30.7}\ / \ \ecs{33.7}\ / \ \fps{32.2} \\
\textbf{GLM-4.6}        & \fones{35.5}\ / \ \ecs{31.5}\ / \ \fps{42.4} & \fones{38.7}\ / \ \ecs{39.9}\ / \ \fps{39.8} & \fones{26.1}\ / \ \ecs{23.1}\ / \ \fps{32.1} & \fones{38.1}\ / \ \ecs{37.3}\ / \ \fps{42.6} \\
\textbf{Doubao-1.6}     & \fones{37.9}\ / \ \ecs{26.5}\ / \ \fps{70.2} & \fones{35.4}\ / \ \ecs{26.3}\ / \ \fps{59.3} & \fones{23.5}\ / \ \ecs{16.7}\ / \ \fps{49.3} & \fones{27.5}\ / \ \ecs{20.4}\ / \ \fps{47.1} \\
\textbf{DeepSeek-V3.1*} & \fones{43.6}\ / \ \ecs{37.2}\ / \ \fps{53.6} & \fones{23.2}\ / \ \ecs{22.4}\ / \ \fps{26.8} & \fones{15.9}\ / \ \ecs{12.8}\ / \ \fps{24.3} & \fones{26.2}\ / \ \ecs{22.2}\ / \ \fps{36.1} \\
\textbf{Kimi-K2*}       & \fones{21.6}\ / \ \ecs{16.0}\ / \ \fps{36.6} & \fones{14.4}\ / \ \ecs{12.1}\ / \ \fps{19.8} & \fones{18.2}\ / \ \ecs{14.0}\ / \ \fps{28.0} & \fones{27.4}\ / \ \ecs{22.5}\ / \ \fps{39.2} \\
\bottomrule
\end{tabular}
\end{adjustbox}

\begin{adjustbox}{width=.85\textwidth}
\begin{tabular}{lcccc}
\toprule
\textbf{Detect $\backslash$ Synth}
& \textbf{Grok-4}
& \textbf{Qwen3-VL}
& \textbf{Doubao-1.6}
& \textbf{GLM-4.6} \\
\midrule
\textbf{Gemini-2.5-Pro} & \fones{34.9}\ / \ \ecs{34.2}\ / \ \fps{36.9} & \fones{38.4}\ / \ \ecs{35.7}\ / \ \fps{43.7} & \fones{47.6}\ / \ \ecs{58.1}\ / \ \fps{41.1} & \fones{30.3}\ / \ \ecs{30.8}\ / \ \fps{30.9} \\
\textbf{GPT-5}          & \fones{33.8}\ / \ \ecs{41.0}\ / \ \fps{30.4} & \fones{38.1}\ / \ \ecs{43.7}\ / \ \fps{35.7} & \fones{42.9}\ / \ \ecs{65.4}\ / \ \fps{33.0} & \fones{29.1}\ / \ \ecs{37.1}\ / \ \fps{25.0} \\
\textbf{Grok-4}         & \fones{28.6}\ / \ \ecs{24.3}\ / \ \fps{37.9} & \fones{31.0}\ / \ \ecs{23.5}\ / \ \fps{50.4} & \fones{47.0}\ / \ \ecs{46.4}\ / \ \fps{50.3} & \fones{23.2}\ / \ \ecs{19.6}\ / \ \fps{33.3} \\
\textbf{Claude-4.5}     & \fones{33.9}\ / \ \ecs{31.4}\ / \ \fps{39.3} & \fones{29.6}\ / \ \ecs{26.6}\ / \ \fps{35.4} & \fones{33.2}\ / \ \ecs{37.2}\ / \ \fps{31.3} & \fones{27.2}\ / \ \ecs{27.6}\ / \ \fps{28.8} \\
\textbf{Qwen3*}         & \fones{29.4}\ / \ \ecs{29.2}\ / \ \fps{31.8} & \fones{31.2}\ / \ \ecs{29.3}\ / \ \fps{34.4} & \fones{35.7}\ / \ \ecs{43.7}\ / \ \fps{31.1} & \fones{25.1}\ / \ \ecs{26.1}\ / \ \fps{25.2} \\
\textbf{O4-Mini}        & \fones{27.7}\ / \ \ecs{28.6}\ / \ \fps{29.7} & \fones{27.4}\ / \ \ecs{25.9}\ / \ \fps{30.6} & \fones{37.8}\ / \ \ecs{48.5}\ / \ \fps{32.8} & \fones{23.2}\ / \ \ecs{24.7}\ / \ \fps{23.4} \\
\textbf{GLM-4.6}        & \fones{29.6}\ / \ \ecs{28.8}\ / \ \fps{31.5} & \fones{27.0}\ / \ \ecs{24.0}\ / \ \fps{33.1} & \fones{33.6}\ / \ \ecs{36.8}\ / \ \fps{31.9} & \fones{25.3}\ / \ \ecs{23.8}\ / \ \fps{28.0} \\
\textbf{Doubao-1.6}     & \fones{21.7}\ / \ \ecs{15.1}\ / \ \fps{43.0} & \fones{19.4}\ / \ \ecs{13.1}\ / \ \fps{42.3} & \fones{42.0}\ / \ \ecs{32.8}\ / \ \fps{62.6} & \fones{13.3}\ / \ \ecs{9.5}\ / \ \fps{24.6} \\
\textbf{DeepSeek-V3.1*} & \fones{19.2}\ / \ \ecs{15.2}\ / \ \fps{30.9} & \fones{21.7}\ / \ \ecs{17.1}\ / \ \fps{33.4} & \fones{20.2}\ / \ \ecs{20.3}\ / \ \fps{21.0} & \fones{19.0}\ / \ \ecs{16.8}\ / \ \fps{24.1} \\
\textbf{Kimi-K2*}       & \fones{20.9}\ / \ \ecs{15.8}\ / \ \fps{33.1} & \fones{15.4}\ / \ \ecs{11.1}\ / \ \fps{27.4} & \fones{16.4}\ / \ \ecs{14.6}\ / \ \fps{19.9} & \fones{15.9}\ / \ \ecs{12.9}\ / \ \fps{23.0} \\
\bottomrule
\end{tabular}
\end{adjustbox}

\caption{Fast-mode detection performance by detection and synthesis models.
Each entry reports \fones{Macro-F1}\ / \ \ecs{Error Coverage (EC)}\ / \ \fps{Finding Precision (FP)} (\%).
Models marked with * denote non-multimodal settings.}
\label{tab:fast_mode_split}
\end{table*}

\subsection{Additional Detectability Analysis}
\label{sec:appendix_detectability}

\paragraph{Detectability by Error Type}
Table~\ref{tab:appendix_error_type_coverage} reports error coverage across different corruption types under various detection settings.
Across all model combinations, increasing analytical depth consistently improves coverage for most error categories.
However, detectability varies substantially by error type.
Errors involving exaggerated claims or rhetorical manipulation, such as Claim Distortion (CD) and Rhetoric Bias (RB), exhibit systematically lower coverage, reflecting the difficulty of identifying subtle over-claiming without explicit contradictions.
In contrast, Context Misalignment (CM), Method Logic (ML), and Ethical Omission (EO) achieve higher coverage, as these errors often involve concrete inconsistencies, invalid reasoning steps, or missing disclosures.
Notably, removing global memory leads to a consistent drop in coverage, particularly for cross-section consistency errors (CS*), underscoring the role of document-level context in reliable error detection.

\begin{table*}[t]
\centering
\scriptsize
\setlength{\tabcolsep}{3pt}  
\begin{adjustbox}{width=.9\textwidth}
\begin{tabular}{l l l c c c c c c c c c c}
\toprule
\textbf{Detect} & \textbf{Synth} & \textbf{Mode} &
\textbf{Overall} & \textbf{CD} & \textbf{CM} & \textbf{EO} & \textbf{EM} & \textbf{EP} & \textbf{ML} & \textbf{RF} & \textbf{RB} & \textbf{CS*} \\
\midrule
\multirow{4}{*}{\textbf{\textit{\shortstack[l]{Claude-\\Sonnet-4.5}}}}
& \multirow{4}{*}{\textbf{\textit{O4-Mini}}}
& Fast                & 35.1 & 30.2 & 56.1 & 37.1 & 28.6 & 25.3 & 34.0 & 32.4 & 46.2 & 53.6 \\
& & Standard          & 48.4 & 44.8 & 62.1 & 50.0 & 40.3 & 37.3 & 46.2 & 58.1 & 52.6 & 56.5 \\
& & \ + \textit{w/o memory} & 45.8 & 40.6 & 47.0 & 54.8 & 40.3 & 46.7 & 45.3 & 50.5 & 46.2 & 46.4 \\
& & Deep              & 57.3 & 44.8 & 72.7 & 59.7 & 48.7 & 46.7 & 61.3 & 67.6 & 61.5 & 65.2 \\
\midrule
\multirow{4}{*}{\textbf{\textit{\shortstack[l]{Gemini-\\2.5-Pro}}}}
& \multirow{4}{*}{\textbf{\textit{GPT-5}}}
& Fast                & 53.8 & 48.9 & 51.6 & 56.2 & 54.4 & 54.5 & 58.3 & 57.4 & 42.0 & 50.0 \\
& & Standard          & 60.7 & 51.1 & 60.4 & 67.2 & 60.8 & 54.5 & 66.5 & 69.3 & 55.1 & 58.3 \\
& & \ + \textit{w/o memory} & 55.4 & 42.3 & 51.6 & 67.2 & 52.6 & 50.3 & 62.4 & 70.3 & 49.3 & 51.4 \\
& & Deep              & 70.3 & 62.0 & 67.0 & 78.1 & 69.0 & 70.6 & 74.3 & 80.2 & 59.4 & 64.7 \\
\midrule
\multirow{4}{*}{\textbf{\textit{GPT-5}}}
& \multirow{4}{*}{\textbf{\textit{GPT-5}}}
& Fast                & 67.7 & 59.1 & 64.8 & 60.9 & 69.0 & 76.2 & 78.0 & 65.3 & 44.9 & 64.0  \\
& & Standard          & 69.6 & 54.0 & 65.9 & 82.8 & 63.2 & 65.7 & 82.6 & 84.2 & 55.1 & 66.5 \\
& & \ + \textit{w/o memory} & 65.3 & 50.4 & 62.6 & 78.1 & 62.0 & 60.1 & 73.9 & 82.2 & 53.6 & 61.2 \\
& & Deep              & 80.6 & 62.0 & 75.8 & 89.1 & 78.4 & 86.0 & 89.4 & 93.1 & 63.8 & 73.7 \\
\bottomrule
\end{tabular}
\end{adjustbox}
\caption{Error Coverage (\%) across different error types under various detection settings.}
\label{tab:appendix_error_type_coverage}
\end{table*}

\subsection{Metric for Human-Agent Alignment}
\label{sec:human_alignment_metrics}

This appendix provides detailed definitions of the ranking-based metrics used to evaluate alignment between automated reviewers and human judgments, complementing the summarized results reported in Section~\ref{sec:alignment_human}.

\paragraph{Human and Model Ranking Keys.}
For each paper, human judgment is represented by a ranking key
$
k_h = (\mu, -\sigma^2),
$
where $\mu$ is the mean overall score across all human reviewers and $\sigma^2$ is the sample variance.
When mean scores are tied, papers with lower inter-reviewer variance are ranked higher, reflecting greater reviewer consensus.

For automated reviewers, we construct a lexicographic ranking key:
$
k_m = (\textit{overall}, \textit{novelty}, \textit{technical\_quality}, \textit{clarity}),
$
where higher values indicate better assessments.
When overall scores are tied, auxiliary dimensions are used sequentially to break ties.
Missing sub-scores are treated as $-\infty$, ensuring they do not win tie-breaks.

\paragraph{Percentile Rank Normalization.}
Given a set of $N$ papers with ranking keys $\{k_i\}_{i=1}^N$, we first sort them in ascending order and assign average ranks to tied items.
Each paper is then mapped to a percentile rank
$
r_i = \frac{\mathrm{rank}(k_i)}{N-1} \in [0,1],
$
where higher values correspond to better-ranked papers.
This normalization ensures comparability across datasets and avoids assumptions about score scales.

\paragraph{Rank-MSE.}
Rank Mean Squared Error (Rank-MSE) measures the squared deviation between human and model percentile ranks:
\[
\text{Rank-MSE} = \frac{1}{N} \sum_{i=1}^N (r_i^{(m)} - r_i^{(h)})^2,
\]
where $r_i^{(m)}$ and $r_i^{(h)}$ denote the model and human percentile ranks, respectively.
Lower values indicate closer alignment between rankings.

\paragraph{Spearman's $\rho$.}
Spearman's rank correlation coefficient $\rho$ is computed as the Pearson correlation between human and model percentile ranks:
\[
\rho = \mathrm{corr}\big(r^{(m)}, r^{(h)}\big).
\]
This metric captures global monotonic consistency between rankings and is insensitive to absolute score differences.

\paragraph{Kendall's $\tau$-b.}
Kendall's $\tau$-b evaluates pairwise ordering agreement while explicitly accounting for ties.
Let $C$ and $D$ denote the numbers of concordant and discordant paper pairs, and let $T_h$ and $T_m$ denote pairs tied only in human or model rankings, respectively.
The $\tau$-b coefficient is defined as
\[
\tau_b = \frac{C - D}{\sqrt{(C + D + T_h)(C + D + T_m)}}.
\]
Values range from $-1$ (complete disagreement) to $1$ (perfect agreement), with $0$ indicating no rank correlation.

\paragraph{Pairwise Accuracy.}
Pairwise Accuracy measures the proportion of concordant pairs among all comparable (non-tied) paper pairs:
\[
\text{Pairwise Accuracy} = \frac{C}{C + D}.
\]
This metric has an intuitive interpretation as the probability that the automated reviewer orders a randomly chosen pair of papers in the same direction as human reviewers.

\paragraph{Interpretation.}
Together, these metrics provide complementary perspectives on alignment.
Rank-MSE emphasizes absolute deviations in normalized rank positions, Spearman's $\rho$ captures global monotonic trends, Kendall's $\tau$-b focuses on pairwise consistency under ties, and Pairwise Accuracy offers an interpretable agreement rate.
All metrics operate purely on relative rankings, avoiding reliance on absolute score calibration, which is known to vary substantially across both human and automated reviewers.

\subsection{Coverage-based Alignment Protocol}
\label{sec:coverage_alignment_prompt}

For coverage-based alignment, we collect real human reviews from ICLR~2026 submissions and first consolidate multiple reviewer comments into a single unified human review summary.
This consolidated summary serves as the reference representation of human judgment.
The AI-generated review is then compared against this human summary using the coverage-based alignment prompt shown in Table~\ref{tab:appendix_prompt_coverage_alignment}.

To ensure a fair comparison, when an automated reviewer (e.g., \textit{DeepReview}) produces multiple reviewer-style outputs, we apply the same consolidation procedure to merge them into a single unified AI review before evaluation.
This normalization step ensures that coverage alignment reflects differences in \emph{content coverage} rather than artifacts of reviewer multiplicity.

\noindent\textbf{Coverage-based Alignment Metrics.}
Let $\mathcal{S}_H$ and $\mathcal{W}_H$ denote the sets of strength points and weakness points extracted from the human review, and $\mathcal{S}_A$ and $\mathcal{W}_A$ denote the corresponding sets extracted from the AI review.
Each point is treated as an atomic, de-duplicated semantic unit.

\paragraph{Strength Coverage Recall (Human $\rightarrow$ AI).}
This metric measures the extent to which the AI review covers the strengths identified by human reviewers:
$$
\text{Str-Cov} = \frac{|\mathcal{S}_H \cap \mathcal{S}_A|}{|\mathcal{S}_H|}.
$$
It reflects how well the AI reviewer captures the core contributions and positive aspects emphasized by humans.

\paragraph{Weakness Coverage Recall (Human $\rightarrow$ AI).}
This metric measures the extent to which the AI review covers the weaknesses or concerns raised by human reviewers:
$$
\text{Weak-Cov} = \frac{|\mathcal{W}_H \cap \mathcal{W}_A|}{|\mathcal{W}_H|}.
$$
It captures the AI reviewer’s ability to attend to the same limitations, methodological issues, or missing analyses highlighted by humans.

\paragraph{AI Extra Major Points Rate.}
This metric quantifies the proportion of major points introduced by the AI review that are not mentioned in the human review.
Let $\mathcal{M}_A = \mathcal{S}_A \cup \mathcal{W}_A$ denote the set of major points emphasized by the AI reviewer.
The metric is defined as:
$$
\text{AI-Extra} = \frac{|\mathcal{M}_A \setminus (\mathcal{S}_H \cup \mathcal{W}_H)|}{|\mathcal{M}_A|}.
$$
Additional points are not necessarily incorrect; rather, this metric reflects divergence in coverage scope.
Lower values indicate closer alignment with human-prioritized issues.

\paragraph{Symmetric Coverage Similarity.}
This metric provides a holistic, symmetric measure of content-level agreement between human and AI reviews.
Let $\mathcal{U} = \mathcal{S}_H \cup \mathcal{W}_H \cup \mathcal{S}_A \cup \mathcal{W}_A$ denote the union of all points, and
$\mathcal{I} = (\mathcal{S}_H \cap \mathcal{S}_A) \cup (\mathcal{W}_H \cap \mathcal{W}_A)$ denote the matched points.
The symmetric coverage similarity is computed as:
$$
\text{Sym-Cov} = \frac{|\mathcal{I}|}{|\mathcal{U}|}.
$$
This metric summarizes overall overlap while penalizing both missed human points and excessive divergence introduced by the AI review.

\subsection{Additional Q2 Review Results}
\label{sec:appendix_q2_additional}

\begin{figure*}[t]
\centering
\includegraphics[width=0.85\textwidth]{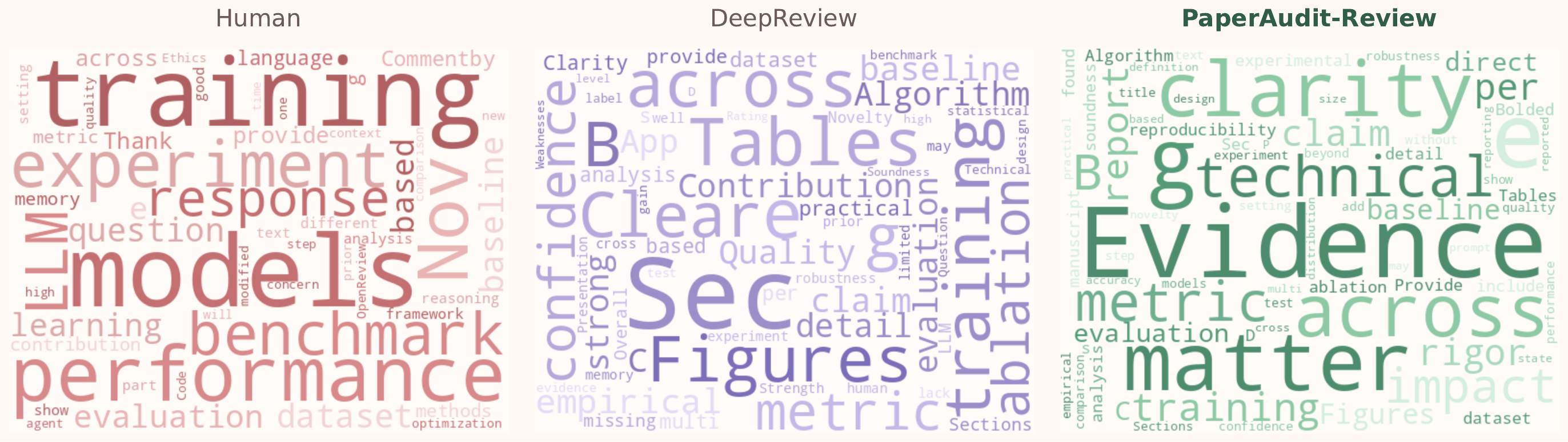}
\caption{Word clouds of review texts under different reviewing methods in Q2. \textit{PaperAudit-Review} places greater emphasis on evidence- and critique-related terms, reflecting a more evidence-grounded reviewing style.}
\label{fig:review_wordclouds}
\end{figure*}

The full numerical review scores and lexical comparison for Q2 are provided below. Table~\ref{tab:vertical_review_comparison} reports the scores, while Figure~\ref{fig:review_wordclouds} visualizes the lexical comparison. Together, these results support the main-text finding that explicit error detection makes automated reviews more critical and evidence-oriented.

\begin{table}[t]
\renewcommand{\arraystretch}{0.95}
\centering
\setlength{\tabcolsep}{3pt}
\begin{adjustbox}{width=.92\columnwidth}
\begin{tabular}{llcccc}
\toprule
\textbf{Reviewer} & \textbf{Synth} & \textbf{Novelty} & \textbf{Tech} & \textbf{Clarity} & \textbf{Overall} \\
\midrule
\multicolumn{6}{l}{\textbf{Baseline-Review}} \\
\midrule
\multirow{3}{*}{\textbf{\textit{\shortstack[l]{Gemini-\\2.5-Pro}}}}
& \textit{\textbf{Origin}}  & 8.70 & 8.30 & 8.09 & 8.20 \\
& \textit{\textbf{GPT-5}}   & 8.70 & 7.40 & 6.96 & 7.78 \\
& \textit{\textbf{O4-Mini}} & 8.52 & 7.43 & 6.96 & 7.48 \\
\midrule
\multirow{3}{*}{\textbf{GPT-5}}
& \textit{\textbf{Origin}}  & 7.61 & 6.63 & 7.19 & 7.27 \\
& \textit{\textbf{GPT-5}}   & 7.30 & 5.75 & 6.17 & 6.63 \\
& \textit{\textbf{O4-Mini}} & 6.97 & 5.79 & 6.09 & 6.41 \\
\midrule
\multicolumn{6}{l}{\textbf{PaperAudit-Review}} \\
\midrule
\multirow{3}{*}{\textbf{\textit{\shortstack[l]{Gemini-\\2.5-Pro}}}}
& \textit{\textbf{Origin}}  & 8.51 & 6.31 & 6.51 & 6.81 \\
& \textit{\textbf{GPT-5}}   & 8.48 & 5.55 & 5.45 & 6.31 \\
& \textit{\textbf{O4-Mini}} & 8.43 & 5.73 & 5.69 & 6.07 \\
\midrule
\multirow{3}{*}{\textbf{\textit{GPT-5}}}
& \textit{\textbf{Origin}}  & 7.61 & 6.15 & 6.55 & 6.93 \\
& \textit{\textbf{GPT-5}}   & 7.29 & 5.45 & 5.89 & 6.45 \\
& \textit{\textbf{O4-Mini}} & 6.99 & 5.55 & 5.76 & 6.28 \\
\midrule
\multicolumn{6}{l}{\textbf{DeepReview \cite{zhu2025deepreview}}} \\
\midrule
\multirow{3}{*}{\textbf{\textit{\shortstack[l]{Gemini-\\2.5-Pro}}}}
& \textit{\textbf{Origin}}  & 7.90 & 7.68 & 7.75 & 7.53 \\
& \textit{\textbf{GPT-5}}   & 7.71 & 6.63 & 7.21 & 6.82 \\
& \textit{\textbf{O4-Mini}} & 7.74 & 6.70 & 6.61 & 6.57 \\
\midrule
\multirow{3}{*}{\textit{\textbf{GPT-5}}}
& \textit{\textbf{Origin}}  & 7.59 & 6.91 & 7.64 & 7.35 \\
& \textit{\textbf{GPT-5}}   & 7.39 & 6.07 & 6.74 & 6.65 \\
& \textit{\textbf{O4-Mini}} & 7.12 & 6.07 & 6.55 & 6.54 \\
\bottomrule
\end{tabular}
\end{adjustbox}
\caption{Average review scores on ICML branch of PaperAudit-Dataset.}
\label{tab:vertical_review_comparison}
\end{table}

\subsection{Error Categories Associated with Review Score Drops}
\label{sec:appendix_review_regression}

\begin{table}[t]
\centering
\small
\setlength{\tabcolsep}{3pt}

\begin{subtable}[t]{\columnwidth}
\centering
\begin{adjustbox}{width=\columnwidth}
\begin{tabular}{lcccc}
\toprule
\textbf{Error Type} & \textbf{\#Drop} & \textbf{Drop Rate} & \textbf{Lift} & \textbf{\#Total} \\
\midrule
Method Logic Errors        & 9  & 0.26 & 1.52 & 35 \\
Context Misalignment      & 8  & 0.22 & 1.25 & 36 \\
Claim Distortion          & 5  & 0.19 & 1.00 & 27 \\
Rhetoric Bias             & 2  & 0.10 & 0.46 & 21 \\
Experiment Protocol Flaws & 1  & 0.08 & 0.40 & 12 \\
Evidence Manipulation     & 1  & 0.12 & 0.63 & 8  \\
\bottomrule
\end{tabular}
\end{adjustbox}
\caption{\textbf{ICML 2025 (synthetic errors with GPT-5).}}
\end{subtable}

\vspace{4pt}

\begin{subtable}[t]{\columnwidth}
\centering
\begin{adjustbox}{width=\columnwidth}
\begin{tabular}{lcccc}
\toprule
\textbf{Error Type} & \textbf{\#Drop} & \textbf{Drop Rate} & \textbf{Lift} & \textbf{\#Total} \\
\midrule
Claim Distortion          & 5  & 0.33 & 1.88 & 15 \\
Method Logic Errors       & 11 & 0.31 & 1.65 & 36 \\
Context Misalignment      & 4  & 0.15 & 0.68 & 26 \\
Rhetoric Bias             & 3  & 0.16 & 0.71 & 19 \\
Experiment Protocol Flaws & 2  & 0.14 & 0.63 & 14 \\
Evidence Manipulation     & 0  & 0.00 & 0.00 & 9  \\
\bottomrule
\end{tabular}
\end{adjustbox}
\caption{\textbf{ICLR 2026 (real submissions with GPT-5 reviewers).}}
\end{subtable}

\caption{\textbf{Error categories most strongly associated with review score drops.}
Top: analysis on GPT-5–synthesized errors for ICML~2025 papers.
Bottom: regression analysis on real GPT-5 reviews for ICLR~2026 submissions.
Lift $>$ 1 indicates a higher likelihood of score reduction relative to non-drop cases.}
\label{tab:review_regression_combined}
\end{table}

To identify which detected issues most strongly contribute to numerical score reductions, we conduct a regression-style analysis that relates changes in review scores to newly surfaced or intensified critiques categorized by PaperAudit error types.
For each paper, we compare baseline and final reviews produced by GPT-5 and examine how the presence of different error categories correlates with subsequent score decreases.
Table~\ref{tab:review_regression_combined} summarizes the results under two complementary settings: GPT-5 reviews on ICML~2025 papers with synthetically injected errors (top), and real GPT-5 reviews of ICLR~2026 submissions (bottom).
For each error category, we report the number of score drops, the empirical drop rate, and a relative \emph{lift}, defined as the ratio between the probability of observing that error in score-decreasing reviews and its probability in non-decreasing reviews. A lift value greater than one therefore indicates that the error type is more strongly associated with score reductions.

Across both settings, error categories that directly affect technical soundness, most notably \textit{method logic errors} and \textit{claim or result distortion}, consistently exhibit higher drop rates and lift values, indicating a stronger influence on scoring decisions.
In contrast, issues related to rhetorical presentation, experimental protocol details, or evidence handling occur frequently but show substantially weaker associations with numerical score decreases.
Overall, this analysis suggests that review score reductions are driven not by the sheer number of detected issues, but by whether those issues undermine methodological consistency or the credibility of reported claims, reinforcing our main conclusion that PaperAudit-Review enables more discriminative and substantively grounded reviewing behavior.

\section{Detailed Pipeline of PaperAudit-Bench}
\subsection{Origin Data Preprocessing}
\label{sec:appendix_data_construction}

We provide \emph{implementation-level details} of the preprocessing pipeline used to
transform raw conference paper PDFs into the structured, section-aware JSON representations
described in Section~\ref{sec:dataset_source}.
We focus on the concrete engineering steps for parsing, cleaning, and organizing original papers
prior to synthetic error injection and auditing.

\paragraph{Paper Acquisition and Length Filtering.}
Source papers are collected from OpenReview through official conference APIs.
For each accepted paper, we create an isolated directory containing the PDF file and associated metadata.
This directory-level isolation is preserved throughout the pipeline to support independent parsing,
failure recovery, and reliable alignment between original and corrupted versions.
Before parsing, we apply a lightweight page-count filter to exclude overly long PDFs, which empirically
lead to unstable OCR or layout extraction due to large appendices or image-heavy content.
Rather than truncating documents, such cases are discarded to ensure consistent and reliable downstream processing.

\paragraph{Multimodal Parsing and Content Extraction.}
Following \cite{son2025ai}, each retained PDF is converted into a structured multimodal representation
using LlamaParse\footnote{\url{https://cloud.llamaindex.ai/}}.
Text is extracted at the page level and refined through a second-stage OCR correction step,
where a language model repairs common layout artifacts (e.g., broken formulas, hyphenation, and line wrapping)
while preserving mathematical expressions and LaTeX-style notation.
In parallel, visual elements such as figures and tables are extracted.
Redundant sub-images and page-level screenshots are removed via bounding-box containment checks
and template-based matching, and remaining candidates are filtered using a lightweight classifier
to distinguish manuscript figures from non-content artifacts.
Validated visual blocks are then interleaved with text blocks to form a unified content sequence.

\paragraph{Section Labeling and Structural Refinement.}
To support section-aware reasoning, parsed text blocks are normalized and segmented with respect to
semantic section boundaries, avoiding arbitrary fragmentation within the same section.
Each content block (text or figure) is assigned a global index and a section label from a fixed predefined set.
Section labels are represented as contiguous index ranges to guarantee full document coverage.
Figures and tables are labeled independently based on captions and semantic cues, allowing their assigned
section to differ from surrounding text when appropriate.

\paragraph{Rule-based Cleanup and Final Representation.}
We apply additional rule-based post-processing to remove systematic noise irrelevant to paper auditing.
Repeated page headers and footers are detected via frequency analysis across blocks and removed while
preserving their first occurrence.
Conference checklist sections are also discarded, as they are orthogonal to scientific claims and
methodological reasoning.
After cleanup, block indices are re-normalized and section spans are rebuilt to maintain internal consistency.

The final output is a section-aware JSON representation in which each paper is stored as an ordered
sequence of indexed multimodal content blocks, serving as the standardized input for error injection,
detection, and review experiments.
Table~\ref{tab:appendix_content_example_with_image} provides an illustrative snippet of the resulting representation.

\begin{table*}[t]
\centering
\small
\textbf{Example: processed paper JSON representation.}
\begin{minipage}{0.93\textwidth}
\begin{verbatim}
"content": [
  {
    "type": "text",
    "text": "Mean Flows for One-step Generative Modeling\n
            Zhengyang Geng ... Kaiming He ...",
    "index": 1,
    "section": "Abstract"
  },
  {
    "type": "text",
    "text": "Abstract\nWe propose a principled and effective framework for
            one-step generative modeling ... (truncated)",
    "index": 2,
    "section": "Abstract"
  },
  {
    "type": "image_url",
    "image_url": "data:image/jpeg;base64,/9j/4AAQSkZJRgABAQAAAQABAAD...
                 ... (truncated)",
    "index": 3,
    "section": "Experiments"
  },
  {
    "type": "text",
    "text": "1 Introduction\nThe goal of generative modeling is to transform
            a prior distribution into the data distribution ... (truncated)",
    "index": 4,
    "section": "Introduction"
  },
  ...
]
\end{verbatim}
\end{minipage}
\caption{Illustrative snippet of a processed paper representation. The \texttt{content} list interleaves section-aware text blocks with image blocks (encoded as base64 \texttt{data:} URLs), and assigns a global index to each block for precise downstream localization.}
\label{tab:appendix_content_example_with_image}
\end{table*}

\subsection{Synthetic Error Injection}
\label{sec:appendix_error_injection}

This subsection summarizes how we inject synthetic errors on top of the processed, section-aware JSON representation.
Concretely, we query a LLM to generate a set of edit patches under strict JSON constraints (Table~\ref{tab:appendix_prompt_injection}), and then deterministically apply these patches to the processed paper content.
Each injected corruption is assigned to one of the eight error categories in Table~\ref{tab:appendix_error_types}.

\paragraph{Edit-based Corruption as JSON Patches.}
We model corruption as a set of \emph{edit patches} applied to the processed paper content.
Given the paper represented as an ordered list of multimodal blocks, the LLM is prompted to propose 10--20 localized edits, where each edit specifies (i) a verbatim text span to locate (\texttt{target\_find}) and (ii) its replacement (\texttt{replacement}), together with lightweight metadata such as the error category (\{Error\_Type\}), the target section, and a short reviewer-style explanation (Table~\ref{tab:appendix_prompt_injection}).
For each modification, we maintain a complete and traceable record, including the original text span, the edited version, the associated category, and a natural-language explanation.
To ensure that edits are precisely grounded and reproducible, \texttt{target\_find} is required to be copied character-for-character from the original text blocks; we then apply the corruption by performing a single-span replacement at the first matched occurrence.
We generate patches with broad section coverage (at least four sections) so that injected errors are not concentrated in a single region, better reflecting diverse failure modes that may appear in real reviewing.

\paragraph{Section-aware Prompting over Block Streams.}
The injector prompts the LLM with the paper represented as an ordered stream of multimodal blocks, each annotated with an index and section label as anchors (Table~\ref{tab:appendix_content_example_with_image}).
Text blocks are provided verbatim (with truncation when needed), while figures and tables are included as \texttt{image\_url} items (base64 \texttt{data:} URLs) to support visually grounded corruptions.
This formulation avoids collapsing the entire PDF into a single long context while preserving global order and section locality, enabling controlled within- and cross-section corruptions.
Each injected patch is finally labeled with an error category (\{Error\_Type\}) from the taxonomy in Table~\ref{tab:appendix_error_types}, spanning evidence manipulation, method logic, protocol flaws, claim distortion, context misalignment, reference fabrication, ethical omission, and rhetoric bias.

\begin{table*}[t]
\centering
\small
\setlength{\tabcolsep}{6pt}
\renewcommand{\arraystretch}{1.15}
\begin{tabular}{p{3.7cm} p{11.2cm}}
\toprule
\textbf{Error Type} & \textbf{Description} \\
\midrule
Evidence Manipulation (EM) &
Incorrect or misleading use of empirical evidence, including fabricated results, altered metrics, or unsupported performance claims that do not faithfully reflect the reported data. \\

Method Logic Errors (ML) &
Logical inconsistencies or invalid reasoning in method or algorithm descriptions, such as incorrect derivations, unjustified assumptions, or flawed step-to-step transitions. \\

Experiment Protocol Flaws (EP) &
Errors in experimental design or evaluation protocols, including invalid baselines, improper ablations, inconsistent evaluation settings, or misleading reporting of experimental procedures. \\

Claim Distortion (CD) &
Overstated, unsupported, or incorrect interpretations of experimental results, where conclusions do not logically follow from the presented evidence or fail under reasonable perturbations. \\

Context Misalignment (CM) &
Inconsistencies across different sections of a paper, such as mismatches between methods, experimental setups, and reported results, often requiring cross-section reasoning to detect. \\

Reference Fabrication (RF) &
Incorrect, misleading, or fabricated citations, including references that do not support the associated claims or formal arguments that misrepresent prior work. \\

Ethical Omission (EO)&
Omission or incomplete disclosure of ethical considerations, such as unreported computational resources, missing risk discussions, or incomplete experimental disclosures. \\

Rhetoric Bias (RB) &
Misleading rhetorical strategies that exaggerate novelty or effectiveness, including biased phrasing, unqualified superlatives, or claims of dominance without sufficient justification. \\
\bottomrule
\end{tabular}
\caption{Detailed descriptions of the eight error categories used in PaperAudit-Dataset.}
\label{tab:appendix_error_types}
\end{table*}

\subsection{Error Detection Pipeline}
\label{sec:appendix_error_detection}
This subsection provides implementation-level and formal details of the error detection workflow summarized in Section~\ref{sec:detect_workflow}. Given a synthesized paper represented as a section-aware block stream (Section~\ref{sec:appendix_error_injection}), the detector outputs a unified list of reviewer-style findings under a strict JSON schema.

The pipeline consists of three complementary detection modules corresponding to the \textit{Fast}, \textit{Standard}, and \textit{Deep} modes described in the main paper: \textbf{(i) global cross-section review}, \textbf{(ii) section-level review with optional global memory}, and \textbf{(iii) task-based multi-agent review}. Each module is designed to expose different classes of issues, ranging from local within-section inconsistencies to cross-section contradictions (often required for error types such as \{Error\_Type\} in Table~\ref{tab:appendix_error_types}).

\paragraph{Fast Mode: Global Cross-section Review.}
In the Fast mode, a document-level detection agent directly operates on the full paper representation $P$ in a single pass, without explicit section decomposition or multi-step reasoning. The detection result is given by
\begin{equation}
\mathcal{F}_{\mathrm{fast}} = f_{\mathrm{det}}(P),
\end{equation}
where $f_{\mathrm{det}}(\cdot)$ denotes a full-document reviewer that scans the ordered block stream and outputs a set of findings. The reviewer is explicitly instructed to prioritize \emph{cross-section support mismatches}, such as abstract-level claims not supported by experimental evidence or numerical inconsistencies across text and tables, while remaining concise and non-redundant. The detailed prompt and strict JSON output schema are provided in Table~\ref{tab:appendix_prompt_global_review}.

\paragraph{Standard Mode: Section-level Review with Global Memory.}
The Standard mode performs section-aware detection augmented with an explicit global memory. Given an input paper $P$, a global memory representation
\begin{equation}
m = f_{\mathrm{mem}}(P)
\end{equation}
is first constructed from the full document, encoding high-level information such as the problem formulation, main contributions, experimental setup, and key claims. The paper is then segmented into a set of section slices $\{s_i\}_{i=1}^{N}$, and each section $s_i$ is independently reviewed by a section-level detector conditioned on both the local section content and the shared global memory. The resulting findings are consolidated by a merging operator:
\begin{equation}
\mathcal{F}_{\mathrm{standard}} = f_{\mathrm{merge}}\Big( \big\{ f_{\mathrm{det}}(s_i, m) \big\}_{i=1}^{N} \Big).
\end{equation}
This design enables fine-grained within-section analysis while supporting cross-section reasoning through shared global context. The corresponding prompt template and unified finding schema are shown in Table~\ref{tab:appendix_prompt_section_review}.

\paragraph{Deep Mode: Task-based Multi-agent Review (Planner $\rightarrow$ Retriever $\rightarrow$ Specialist).}
The Deep mode extends the Standard setting by introducing explicit planning and specialized analysis. For each section $s_i$, a Planner agent generates a targeted review plan
\begin{equation}
\pi_i = f_{\mathrm{plan}}(s_i, m),
\end{equation}
which specifies the review objectives and risk dimensions to be examined. For each plan, a Retriever agent optionally gathers task-specific supporting evidence
\begin{equation}
e_i = f_{\mathrm{ret}}(s_i, m, \pi_i),
\end{equation}
including scholarly title, citation, and arXiv-style lookup when necessary. At evaluation time, most source candidates were unavailable on arXiv, limiting direct retrieval of the original uncorrupted paper, although indirect leakage cannot be ruled out. Guided by $(\pi_i, e_i)$, domain-specific Specialist agents conduct focused reviews and produce findings grounded in the retrieved evidence and local context. The final detection results are obtained by consolidating all specialist outputs:
\begin{equation}
\mathcal{F}_{\mathrm{deep}} = f_{\mathrm{merge}}\Big( \big\{ f_{\mathrm{spec}}(s_i, m, \pi_i, e_i) \big\}_{i=1}^{N} \Big).
\end{equation}
This task-based multi-agent design improves coverage and encourages systematic exploration of diverse error modes, at the cost of higher computational complexity. Prompt templates for the Planner, Retriever, and Specialist roles are provided in Table~\ref{tab:appendix_prompt_task_review}.

\paragraph{Findings Merge and Evaluation.}
Findings produced by the Standard and Deep modes are consolidated through a conservative merge stage that removes exact or near-duplicate reports while preserving distinct, well-supported issues.
This merge step performs adjudication rather than re-detection: it does not introduce new findings or evidence, but only unifies semantically identical reports according to the prompt in Table~\ref{tab:appendix_prompt_merge}.
For evaluation, we adopt an LLM-as-a-Judge protocol to match merged findings against injected ground-truth (GT) errors.
A GT error is considered detected if any finding captures the same underlying issue, allowing paraphrasing and partial span overlap.
Matching is performed conservatively and is not based on exact string matching.
The evaluation prompt and matching criteria are provided in Table~\ref{tab:appendix_prompt_eval_matching}.

\subsection{Training Details of Lightweight Detectors}
\label{sec:appendix_training_details}

\begin{figure}[t]
  \includegraphics[width=\linewidth]{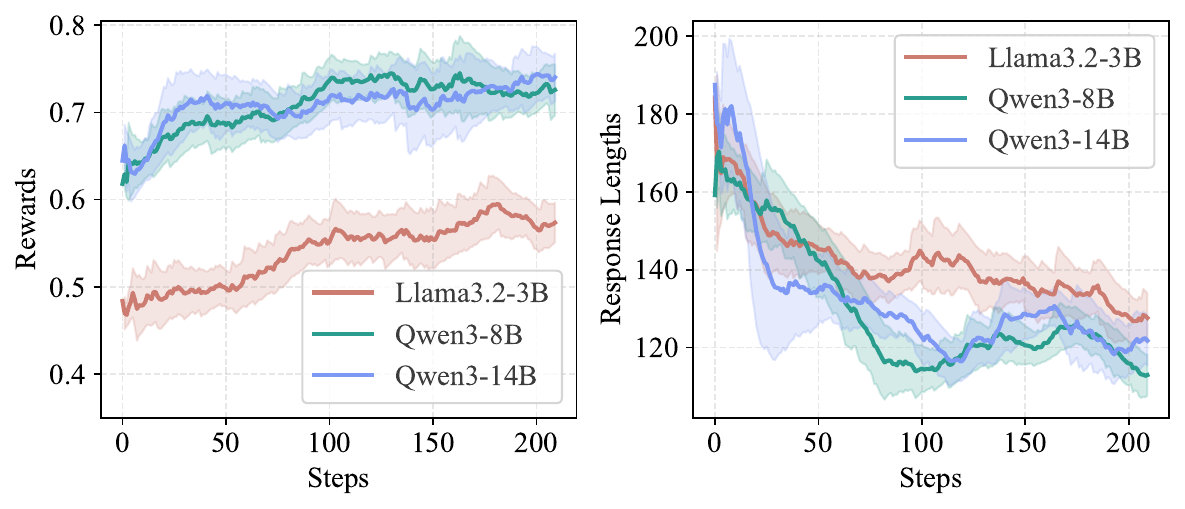}
  \caption{RL training dynamics for lightweight detectors. Reward increases while response length decreases, indicating improved detection quality and output discipline.}
  \label{fig:reward_length}
\end{figure}

For SFT, we use batch size 2, learning rate 5e-5, and 3 epochs with cosine decay and 0.1 warmup ratio.
RL uses rollout batch 2, training batch 64, learning rate 5e-5, cosine decay over 3 epochs, and KL coefficient $\beta=0.3$.
LoRA is applied to all modules with rank $r=64$, $\alpha=128$, and dropout 0.05.
Optimization uses FlashAttention, Adam with CPU offloading, and ZeRO-3.
SFT was conducted on 4 NVIDIA H100 GPUs (0.5h / 2h / 3h for the 3B / 8B / 14B models), while RL used 8 H100 GPUs (2h / 4h / 7h). Qwen3-Next-80B-A3B-Instruct was deployed as the reward model on a separate 4-GPU server because its long-context capacity supports section-level prediction--target comparison without truncating the structured evidence. Table~\ref{tab:training_data_example} gives a representative SFT instance, and Figure~\ref{fig:reward_length} shows the RL training dynamics.

\paragraph{Multi-seed Stability and Reward Sensitivity.}
For Qwen3-1.7B SFT+RL, three seeds yield $30.08\!\pm\!0.55$ EC, $33.56\!\pm\!0.65$ FP, and $31.70\!\pm\!0.62$ F1 under Detection@all. Table~\ref{tab:reward_weight_sensitivity} varies the precision and conciseness weights while holding all other settings fixed. Performance remains broadly comparable across settings, indicating that the detector is not strictly sensitive to the exact weight allocation.

\begin{table*}[t]
\centering
\setlength{\tabcolsep}{4pt}
\renewcommand{\arraystretch}{1.08}
\begin{adjustbox}{max width=\textwidth,center}
\begin{tabular}{lcccccc}
\toprule
\textbf{Weights $(w_p,w_c)$} & \multicolumn{3}{c}{\textbf{Detection@all}} & \multicolumn{3}{c}{\textbf{Detection@1}} \\
\cmidrule(lr){2-4}\cmidrule(lr){5-7}
 & \textbf{EC} & \textbf{FP} & \textbf{F1} & \textbf{EC} & \textbf{FP} & \textbf{F1} \\
\midrule
$(0.2,0.8)$ & \textbf{30.14} & \textbf{33.58} & \textbf{31.77} & 28.03 & 31.21 & 29.53 \\
$(0.4,0.6)$ & 28.58 & 31.84 & 30.13 & 28.73 & 32.01 & 30.28 \\
$(0.6,0.4)$ & 29.77 & 33.17 & 31.38 & \textbf{29.10} & \textbf{32.42} & \textbf{30.67} \\
$(0.8,0.2)$ & 30.07 & 33.50 & 31.69 & 28.94 & 32.26 & 30.51 \\
\bottomrule
\end{tabular}
\end{adjustbox}
\caption{Reward-weight sensitivity for Qwen3-1.7B SFT+RL (\%). Best performance in each evaluation column is bolded.}
\label{tab:reward_weight_sensitivity}
\end{table*}

\begin{table*}[t]
\centering
\setlength{\tabcolsep}{4pt}
\renewcommand{\arraystretch}{1.1}
\begin{adjustbox}{width=0.9\textwidth}
\begin{tabular}{p{\textwidth}}
\toprule
\textbf{Training Instance: Reference Fabrication Detection} \\
\midrule
\textcolor{blue}{\textbf{Input (excerpt):}} \textit{``\ldots \textcolor{firebrick}{KVzip builds on groundbreaking work in [61], which first proposed query-agnostic eviction, making our method a direct extension.} Unlike existing methods \ldots''} \\[3pt]
\textcolor{blue}{\textbf{Output:}} \texttt{\{"error\_type": "\textcolor{darkcyan}{reference\_fabrication}", "error\_location": "\textcolor{firebrick}{KVzip builds on [61]\ldots}", "explanation": "Citation [61] is a RAG paper unrelated to KV compression; this falsely inflates novelty."\}} \\[3pt]
\textbf{Matching Score: 0.88} \\
\bottomrule
\end{tabular}
\end{adjustbox}
\caption{A representative SFT training instance for error detection. The model identifies a \textit{reference fabrication} error where [61] is misattributed to inflate novelty.}
\label{tab:training_data_example}
\end{table*}

\subsection{Review Workflow Details}
\label{sec:appendix_review_prompts}

Following the workflow in Section~\ref{sec:review_pipeline}, we examine how explicit error detection supports more critical peer review.
Observed score differences arise from conservative, prompt-constrained refinements grounded in the manuscript, rather than from free-form re-evaluation.

\paragraph{Baseline Review and Document Memory.}
We first build a high-density \emph{paper memory} from the full manuscript, which serves as a compact, section-aligned recall artifact for downstream analysis (Table~\ref{tab:appendix_prompt_memory}).
This memory is strictly descriptive: it preserves the paper’s stated problem, method, experimental setup, key quantitative results, and explicitly reported limitations, while forbidding critique or extrapolation.
Independently, a baseline reviewer then produces a standard reviewer-style assessment under a fixed structure and evidence-first constraints (Table~\ref{tab:appendix_prompt_baseline}), without access to any auxiliary analyses.

\paragraph{Auxiliary Reports as Structured Analytical Signals.}
Next, we construct two auxiliary reports that function as \emph{silent attention cues} for later refinement, rather than as evidence sources.
\textbf{(i) Error/Integrity Report} is generated by an integrity-focused reviewer that scans the manuscript (and optionally section slices with global memory) to surface \emph{high-impact, clearly observable} issues (Table~\ref{tab:appendix_prompt_error}).
Concretely, it prioritizes: (a) cross-section contradictions (claims vs.\ methods vs.\ results), (b) numerical or factual mismatches across text, figures, tables, and equations, and (c) missing or conflicting protocol details that prevent verification.
The report is constrained to be \emph{evidence-first} and conservative: every issue must be supported by explicit manuscript anchors (e.g., section/figure/table/equation/page), speculative accusations are prohibited, and minor stylistic issues are ignored; when direct support is unavailable, the report must explicitly state that no evidence was found.
\textbf{(ii) Motivation/Novelty Report} is generated by a Senior-AC-style evaluator that cross-examines the authors’ stated motivation and novelty claims against a retrieved set of similar works (Table~\ref{tab:appendix_prompt_motivation}).
This module first extracts the claimed problem, gap, and solution from the manuscript, then evaluates whether the paper explicitly differentiates itself from similar prior works and whether such differentiation is technically substantive.
Crucially, the evaluator is forbidden from hallucinating missing citations or external facts: novelty judgments must cite manuscript sections (e.g., Introduction/Related Work/Discussion) and provide a synthesized verdict (Incremental vs.\ Application-Oriented vs.\ Substantive) that goes beyond merely restating similarity evidence.

\paragraph{Prompt-constrained Refinement.}
Finally, the baseline review is refined under strict minimal-edit constraints (Table~\ref{tab:appendix_prompt_refine}).
The refinement prompt enforces that the baseline review remains the primary artifact: global rewriting, stylistic polishing, and speculative extensions are disallowed.
Although the two auxiliary reports are provided to guide where the reviewer should re-check the manuscript, the reviewer is explicitly forbidden from citing, quoting, or referring to them in any form.
All critiques and score adjustments must be justified exclusively by manuscript evidence with explicit anchors; if no direct evidence can be identified, concerns are not escalated.
Together, these constraints ensure that structured analyses calibrate the review process without replacing it, yielding stricter yet accountable human-style peer review.

\section{Case Study}
\label{sec:appendix_case_study}
In this section, we present two representative case studies to qualitatively illustrate the behavior of our system, drawn from different stages of the pipeline: one focusing on the \textit{error detection agent} and the other on the \textit{review refinement agent}.

\subsection{Error Detection Case}
\label{sec:appendix_case_study_detect}

\paragraph{Detect Example on a Corrupted Paper.}
We present a case study using GPT-5 as both the corruption generator and the detection agent on a corrupted ICML-25 paper (\textit{STAIR: Improving Safety Alignment with Introspective Reasoning}).
As shown in Table~\ref{tab:detect_case_stair_modes}, the three detection modes exhibit complementary behaviors.
The \textit{Fast} mode reliably captures salient surface-level inconsistencies, such as exaggerated claims and narrative--table mismatches that require minimal contextual reasoning.
Errors that depend on cross-section reconciliation or theoretical consistency, including train--test contamination, normalization mismatches, and decoding protocol inconsistencies, are primarily identified under the \textit{Standard} and \textit{Deep} modes.
In contrast, errors that are fundamentally unverifiable from the manuscript alone, such as misreported computation costs or unexplained numerical changes, remain challenging across all modes, reflecting inherent limitations of manuscript-based auditing.

\paragraph{Fine-grained and Over-analysis.}
Deeper detection modes uncover a broader set of fine-grained error signals, but also introduce over-correction on details peripheral to the core scientific claims.
As shown in Table~\ref{tab:deep_unmatched_case_study}, many unmatched findings correspond to valid but out-of-scope analyses, including evaluator coupling, comparability controls, and undocumented evaluation discrepancies.
At the same time, Deep inspection produces over-analytical observations, such as exhaustive reproducibility checklists or presentation-level critiques, which are more appropriately viewed as reviewer-style suggestions than concrete errors.
These results underscore the role of human judgment in interpreting detection outputs, with deeper modes being particularly suited for self-auditing on familiar manuscripts, where authors can selectively adopt substantive corrections while disregarding low-impact over-analysis.

\subsection{Review Case}
\label{sec:appendix_case_study_review}
\paragraph{Review Comparison.}
We present a qualitative review case study on the corrupted paper (\textit{STAIR}) to examine how explicit error detection influences downstream peer review.
Table~\ref{tab:review_case_study_comparison} compares the baseline review with an error-aware \textit{PaperAudit} review produced from the same manuscript.
While the baseline review focuses on conceptual contributions and reported empirical gains, the \textit{PaperAudit} review systematically evaluates internal consistency and evaluation validity.
This leads to the identification of theory inconsistencies, contamination risks, and claim--evidence mismatches that are not raised in the baseline assessment.
Correspondingly, the audit-oriented review assigns more conservative scores in technical quality and clarity, reflecting evidence-based scrutiny rather than stylistic strictness.

\paragraph{Motivation and Error Reports.}
In addition to review-level differences, we also present the associated motivation report and detected error report as supplementary case-study artifacts.
These reports summarize the stated research motivation and enumerate evidence-anchored consistency issues identified from the same corrupted manuscript, as shown in Table~\ref{tab:case3_motivation_integrity_reports}.

\subsection{Deployment on ScienceOne}
\label{sec:appendix_deployment}
\begin{figure*}[t]
  \centering
  \includegraphics[width=.95\linewidth]{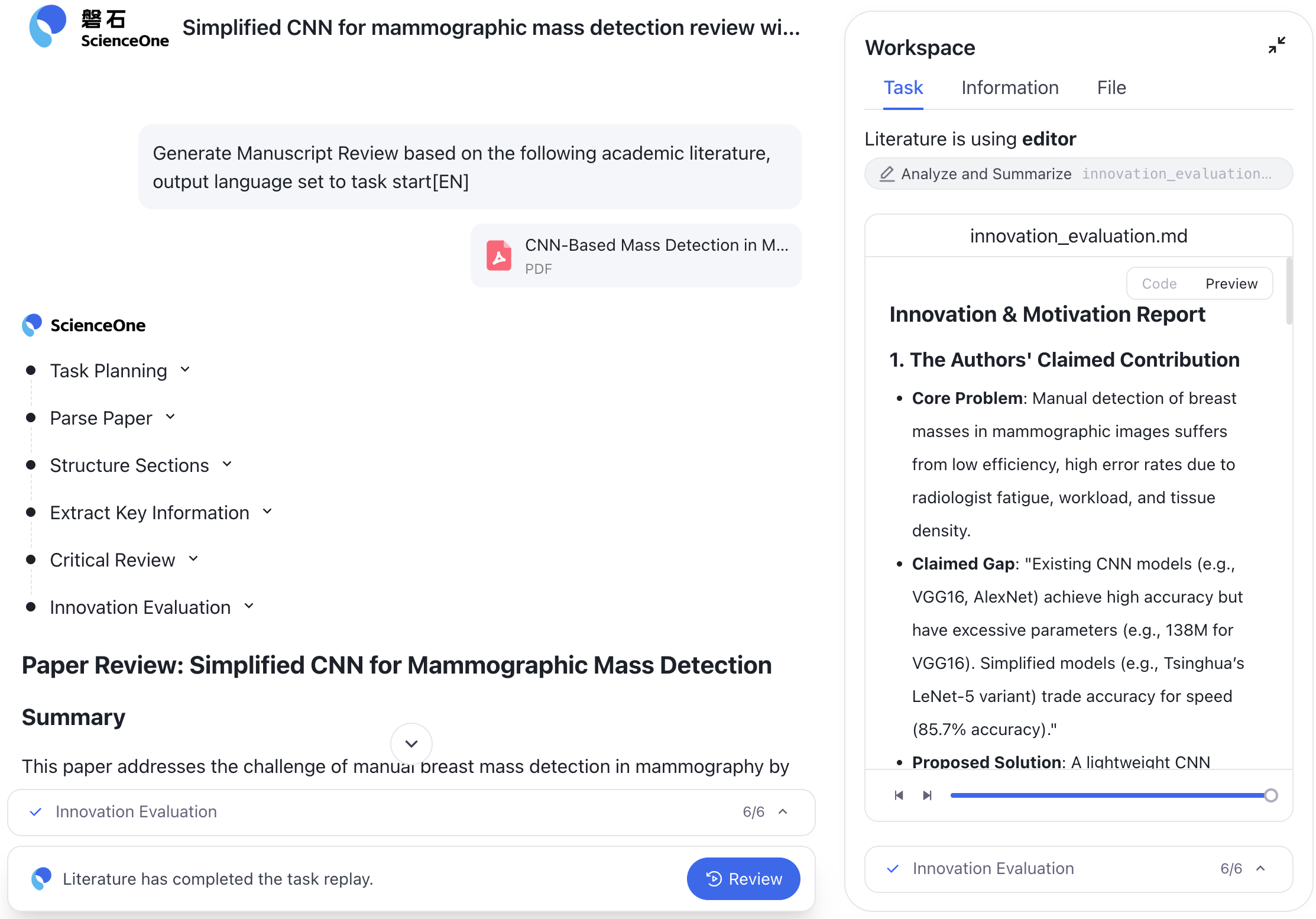}
  \caption{The deployed Manuscript Review module in ScienceOne, a scientific foundation-model platform led by the Chinese Academy of Sciences. The interface shows the PaperAudit-Review workflow from PDF parsing and section structuring to critical review and innovation evaluation.}
  \label{fig:deployment_interface}
\end{figure*}

PaperAudit-Review has been deployed as the Manuscript Review module of \textbf{ScienceOne}, a scientific foundation-model platform led by the Chinese Academy of Sciences.
Given an uploaded PDF, the service performs task planning, paper parsing, section structuring, key-information extraction, critical review, and innovation evaluation before returning a structured academic review.
Figure~\ref{fig:deployment_interface} shows the deployed interface. The deployment demonstrates practical integration as an assistive pre-submission auditing service; it is not intended to replace human peer review. The service is available at \url{https://www.scienceone.cn/lit/\#/assistant}.

\section{Prompts}
Tables~\ref{tab:appendix_prompt_injection}--\ref{tab:appendix_prompt_refine} summarize all prompt templates used in this work.
For clarity and space efficiency, some prompts are lightly condensed, while preserving their original constraints and operational semantics.
Complete prompt implementations and full specifications are available in the released code.


\begin{table*}[t]
\centering
\small
\setlength{\tabcolsep}{6pt}
\renewcommand{\arraystretch}{1.2}
\begin{tabular}{p{0.93\textwidth}}
\toprule
\textbf{Case 2 (Detect Example: Error Coverage across Detection Modes)} \\
\midrule
\textcolor{blue}{\textbf{Legend.}}
\cmark = detected \quad
\xmark = missed \quad
(Columns: \textbf{F}ast / \textbf{S}tandard / \textbf{D}eep) \\[4pt]
\begin{tabular}{@{}p{0.80\textwidth} c c c@{}}
\toprule
\textbf{Injected Error (condensed GT)} & \textbf{F} & \textbf{S} & \textbf{D} \\
\midrule

\textbf{E1 (Abstract).} Over-claim: surpasses Claude-3.5 on all jailbreak suites with no helpfulness loss.
& \cmark & \cmark & \cmark \\

\textbf{E2 (Method).} Inconsistent output delimiters: \texttt{<|Final|>} vs.\ \texttt{<|Output|>} across sections.
& \cmark & \cmark & \cmark \\

\textbf{E3 (Method).} Theorem inconsistency: $F(0)=1$ contradicts Appendix proof requiring $F(0)=0$.
& \cmark & \cmark & \cmark \\

\textbf{E4 (Experiments).} Fabricated narrative number: AlpacaEval 48.66\% vs.\ tables $\sim$38.66\%.
& \cmark & \cmark & \cmark \\

\textbf{E5 (Experiments).} Train--test contamination: StrongReject prompts included in training data.
& \cmark & \cmark & \cmark \\

\textbf{E6 (Appendix).} Ethical impact statement weakens or omits jailbreak risk discussion.
& \xmark & \cmark & \cmark \\

\textbf{E7 (Experiments).} Fabricated benchmark claim: HarmBench-XL listed without results.
& \xmark & \cmark & \cmark \\

\textbf{E8 (References).} Unverifiable citation (Li et al., 2025) used to justify HarmBench-XL.
& \xmark & \xmark & \xmark \\

\textbf{E9 (Appendix).} UCB1 formula inverted ($\ln n_i/N_i$ instead of $\ln N_i/n_i$).
& \cmark & \xmark & \cmark \\

\textbf{E10 (Experiments).} Misreported compute time (30h $\rightarrow$ 8h on A800).
& \xmark & \xmark & \xmark \\

\textbf{E11 (Table).} Arithmetic inconsistency: Table~7 average altered without split changes.
& \xmark & \cmark & \xmark \\

\textbf{E12 (Conclusion).} Over-generalized claim: eliminates jailbreak vulnerability without trade-offs.
& \cmark & \xmark & \cmark \\

\textbf{E13 (Appendix).} Removal of evaluation-set decontamination description.
& \xmark & \cmark & \cmark \\

\textbf{E14 (Appendix).} Default decoding inconsistency: beam search vs.\ greedy.
& \xmark & \xmark & \cmark \\

\textbf{E15 (Table).} Unexplained StrongReject score change (0.6753 $\rightarrow$ 0.7153).
& \xmark & \xmark & \xmark \\

\textbf{E16 (Method+Appendix).} Rating normalization mismatch: $[-1,1]$ vs.\ $[0,1]$.
& \cmark & \cmark & \cmark \\

\bottomrule
\end{tabular}
\end{tabular}

\caption{Detect-agent case study on a corrupted \textit{STAIR} paper. Each row lists an injected paper-level error and whether it is detected under \textit{Fast}, \textit{Standard}, and \textit{Deep} modes. Fast captures prominent surface-level inconsistencies, while Standard and Deep progressively recover cross-section and theory-sensitive errors.}
\label{tab:detect_case_stair_modes}
\end{table*}

\begin{table*}[t]
\centering
\small
\setlength{\tabcolsep}{6pt}
\renewcommand{\arraystretch}{1.2}
\begin{tabular}{p{0.25\linewidth} p{0.65\linewidth}}
\toprule
\multicolumn{2}{l}{\textbf{Case 2 (Detect Example: Fine-grained and Over-analysis)}} \\
\midrule

\multicolumn{2}{l}{\textbf{A. Fine-grained Analysis (Valid but Out-of-scope to Injected GT)}} \\
\midrule

Evaluator coupling
& Identifies that GPT-4o is used both to generate or filter training data and as an evaluator for StrongReject, raising \Fine{potential judge-specific bias}. This is a \Fine{substantive evaluation concern}, but not an injected corruption. \\

Comparability control
& Points out that claims of fair comparison mix \Fine{8B open models with a 32B model}, confounding \Fine{performance attribution}. This reflects experimental rigor rather than a planted error. \\

Cross-table inconsistency
& Notes that STAIR-DPO-3 reports \Fine{different AlpacaEval scores across tables} under seemingly identical settings, suggesting \Fine{undocumented evaluation differences}. This is distinct from the injected narrative-number manipulation. \\

Safety disclosure
& Observes that qualitative examples include \Fine{harmful prompts without sufficient redaction or safety framing} in figures, a legitimate concern about responsible presentation. \\

Conceptual precision
& Questions the \Fine{causal framing} that rapid refusals imply vulnerability, highlighting an \Fine{imprecise conceptual leap} rather than a factual contradiction. \\

\midrule
\multicolumn{2}{l}{\textbf{B. Over-analysis (Reviewer-style or Low-specificity Observations)}} \\
\midrule

Reproducibility granularity
& Requests \Over{exhaustive disclosure of hyperparameters, random seeds, and rollout schedules} beyond what is required to verify core claims. \\

Algorithmic underspecification
& Flags \Over{missing threshold values and stopping criteria} in the algorithm description, which affects clarity but does not constitute a concrete error. \\

Figure hygiene
& Focuses on \Over{missing axis labels or unclear compute scales} in plots, a presentation issue rather than a scientific inconsistency. \\

Bibliographic completeness
& Points out \Over{missing venue details or formatting issues} in references, which impacts polish but not validity. \\

Versioning detail
& Requests a \Over{specific code commit hash or release tag}, reflecting best practices rather than an error in the manuscript. \\

\bottomrule
\end{tabular}
\caption{Case study of representative \emph{unmatched} findings produced by the Deep detection mode.
\Fine{Blue} highlights fine-grained but substantive analyses beyond the injected ground-truth scope, while \Over{orange} marks reviewer-style or low-specificity observations.}
\label{tab:deep_unmatched_case_study}
\end{table*}

\begin{table*}[t]
\centering
\small
\setlength{\tabcolsep}{6pt}
\renewcommand{\arraystretch}{1.15}
\begin{tabular}{p{0.42\linewidth} p{0.5\linewidth}}
\toprule
\multicolumn{2}{l}{\textbf{Case 3 (Review Example: Comparison of Baseline and PaperAudit Review)}} \\
\midrule
\textbf{Baseline Review} & \textbf{PaperAudit Review} \\
\midrule

\textbf{Overall Tone} &
\textbf{Overall Tone} \\
Emphasizes conceptual novelty, strong empirical gains, and clarity of the proposed SI-MCTS + step-level DPO framework. &
\auditcrit{More Critical and Evidence-Driven Tone}.
Focuses on \auditcrit{validity, internal consistency, and evaluation integrity}, even when core ideas are promising. \\

\midrule
\textbf{Strengths Emphasis} &
\textbf{Strengths with Qualification} \\

Highlights dual-process framing, theoretically grounded reward design, and strong StrongReject improvements. &
Acknowledges the same strengths, but \auditwarn{systematically qualifies claims} by linking them to specific assumptions, evaluation setups, and boundary conditions. \\

\midrule
\textbf{Treatment of Theory} &
\textbf{Theory Auditing} \\

Notes the reward formulation and Theorem~2.1 as principled and well-motivated. &
\auditcrit{Explicitly flags the inconsistency between $F(0)=1$ in Section~2.2 and $F(0)=0$ in Appendix~B.1}, questioning correctness of the stated theorem family and its conditions. \\

\midrule
\textbf{Evaluation and Results} &
\textbf{Evaluation Integrity Checks} \\

Accepts reported gains on StrongReject, AlpacaEval, and comparisons to open/proprietary models at face value. &
\auditcrit{Identifies train--test contamination risk} from StrongReject prompts used in training and \auditcrit{numerical inconsistencies across text and tables} (e.g., AlpacaEval 48.66\% vs.\ 38.66\%). \\

\midrule
\textbf{Use of External Models} &
\textbf{Evaluator Coupling and Bias} \\

Mentions GPT-4o usage mainly as an implementation detail for data synthesis and grading. &
\auditwarn{Highlights evaluator coupling}, where GPT-4o is used both for data construction and as the StrongReject evaluator, raising concerns about bias and inflated gains. \\

\midrule
\textbf{Algorithmic Details} &
\textbf{Algorithmic Correctness} \\

Treats SI-MCTS description and UCB-style selection as sufficiently specified. &
\auditcrit{Detects incorrect UCB1 formulation} and missing equation references, directly affecting the validity of Algorithm~1. \\

\midrule
\textbf{Statistical Reporting} &
\textbf{Statistical and Reporting Gaps} \\

Does not question the absence of variance estimates or confidence intervals. &
\auditwarn{Notes missing variance, error bars, and sensitivity analyses}, limiting robustness of empirical conclusions. \\

\midrule
\textbf{Scope and Claims} &
\textbf{Claim--Evidence Alignment} \\

Generally accepts abstract and conclusion claims. &
\auditcrit{Flags over-generalized claims} such as “surpasses every jailbreak suite” and unsupported mentions of HarmBench-XL without reported results. \\

\midrule
\textbf{Final Scores} &
\textbf{Final Scores} \\

Overall: 7 \quad Novelty: 7 \quad Technical: 6 \quad Clarity: 7  &
\auditcrit{Overall: 6} \textbf{\textcolor{eccolor}{$\downarrow$}} \quad Novelty: 7 \quad \auditcrit{Technical: 5} \textbf{\textcolor{eccolor}{$\downarrow$}} \quad \auditcrit{Clarity: 6} \textbf{\textcolor{eccolor}{$\downarrow$}}  \\

\bottomrule
\end{tabular}
\caption{Comparing the baseline review and an \textit{PaperAudit} review on the corrupted  \textit{STAIR} paper.
The audit-oriented review systematically surfaces theory inconsistencies, evaluation integrity risks, and claim--evidence misalignments that are largely unaddressed in the baseline review.
\auditcrit{TealBlue} highlights substantively auditable issues that directly affect correctness or evaluation validity, while \auditwarn{Plum} denotes reviewer-level concerns that are reasonable but not strictly erroneous.}
\label{tab:review_case_study_comparison}
\end{table*}

\begin{table*}[t]
\centering
\small
\setlength{\tabcolsep}{6pt}
\renewcommand{\arraystretch}{1.25}
\begin{tabular}{p{0.93\textwidth}}
\toprule
\textbf{Case 3 (Review Example: Motivation and Detected Error Reports)} \\
\midrule

\textcolor{GoodGreen}{\textbf{Innovation \& Motivation Report}} \\

\good{Targeted Problem.}
The paper targets safety alignment of large language models under jailbreak scenarios, emphasizing the safety--performance trade-off induced by reflexive refusals and shallow alignment strategies. \\

\good{Claimed Gap.}
Conventional SFT/DPO/RLHF approaches are described as effective for overt harms but vulnerable to complex jailbreaks, while existing multi-objective and deliberative alignment methods are argued to be insufficient or to require large reasoning models. \\

\good{Proposed Framework.}
The manuscript introduces a three-stage pipeline consisting of structured chain-of-thought format alignment, iterative self-improvement via Safety-Informed Monte Carlo Tree Search, and a process reward model guiding test-time Best-of-$N$ and beam search. \\

\good{Reward Design.}
A theoretically motivated reward family balancing helpfulness and safety is proposed and instantiated as $R(H,S)=S\!\cdot\!H+2S$, with stated properties such as safety priority and dual monotonicity. \\

\good{Novelty Characterization.}
The contribution is positioned as an integration of known components into a safety-focused, process-level alignment pipeline, with novelty primarily arising from engineering synthesis and targeted application. \\

\midrule
\textcolor{BadRed}{\textbf{Academic Integrity and Consistency Risk Report}} \\

\bad{Theoretical Consistency.}
The reward formulation exhibits inconsistent conditions on $F(0)$ between the main theorem and the appendix, leading to incompatible assumptions for the stated theoretical family and its instantiation. \\

\bad{Reward Scale and Implementation.}
Theoretical assumptions require safety scores in $[-1,1]$ with negative values for unsafe outputs, while implementation descriptions normalize ratings to $[0,1]$ without documenting any re-centering, undermining stated properties. \\

\bad{Search Algorithm Specification.}
The UCB1 selection rule is described with inverted parent--child visit terms relative to the standard formulation, and a referenced equation is missing, affecting algorithmic clarity and verifiability. \\

\bad{Formatting and Data Specification.}
Final-answer and reasoning-step tag formats are inconsistent across sections and appendices, introducing ambiguity in training, parsing, and evaluation procedures. \\

\bad{Numerical Reporting.}
Central metrics such as AlpacaEval winning rates are reported inconsistently across narrative text and tables, with multiple discrepancies lacking documented changes in evaluation settings. \\

\bad{Claim Alignment.}
Abstract-level claims of per-suite superiority and zero degradation in helpfulness are not uniformly supported by the reported per-benchmark results. \\

\bottomrule
\end{tabular}
\caption{Motivation and detected error reports for the corrupted ICML-25 paper \textit{STAIR}.
\good{Green} indicates reported strengths and motivations, while \bad{red} indicates reported inconsistencies and risks, without additional interpretation.}
\label{tab:case3_motivation_integrity_reports}
\end{table*}

\begin{table*}[b]
\centering
\small
\setlength{\tabcolsep}{6pt}
\renewcommand{\arraystretch}{1.15}
\begin{tabular}{p{0.93\textwidth}}
\toprule
\textbf{Prompt Template (Synthetic Error Injection)} \\
\midrule
\textcolor{blue}{\textbf{System Prompt}} \\
You are a strict JSON-only generator. Output a single valid JSON object and nothing else. \\
You are a synthetic corruption generator for academic papers. \\
You will receive the paper as an ordered list of blocks (each block is either a TEXT chunk or an IMAGE URL). \\
Read blocks in order. Do NOT reconstruct the full paper; propose realistic edits based on the provided blocks. \\
\\
\textbf{Error Types} \\
Choose exactly one \texttt{corruption\_type} (\{Error\_Type\}) per edit. Error type definitions follow Table~\ref{tab:appendix_error_types}. \\
\\
\textbf{Edit Coverage Requirements} \\
- Produce 10--20 edits. \\
- Cover at least four sections among: Abstract, Introduction, Related Work, Method, Experiments, Discussion, Conclusion, References. \\
- Keep coherence; edits must resemble genuine manuscript content. \\
\\
\textbf{Output Rules} \\
Return ONE SINGLE JSON object with keys: \\
\texttt{"global\_explanation"} (string) and \texttt{"edits"} (list of 10--20 objects). \\
Each edit object must contain exactly: 
\begin{verbatim}
{
  "id": 1,
  "corruption_type": "{Error_Type}",
  "difficulty": "easy|medium|hard",
  "location": "SectionName",
  "rationale": "short reason",
  "error_explanation": "reviewer-style, 3-6 sentences, with anchors",
  "needs_cross_section": true|false,
  "target_find": "verbatim substring copied from original TEXT blocks",
  "replacement": "new content replacing target_find (concise)"
}
\end{verbatim} 
\\
\\
\textbf{Hard Constraints} \\
- \texttt{target\_find} MUST be copied verbatim from the original TEXT blocks (character-for-character). \\
- Prefer selecting \texttt{target\_find} from TEXT blocks; if image-driven, anchor via nearest caption text. \\
- Do NOT output the full paper or full tables; keep edits paragraph-level and self-contained. \\
\bottomrule
\end{tabular}
\caption{Prompt template used to synthesize document-level corruptions as JSON edit patches. The error type placeholder \{Error\_Type\} refers to the taxonomy in Table~\ref{tab:appendix_error_types}.}
\label{tab:appendix_prompt_injection}
\end{table*}

\begin{table*}[htbp]
\centering
\small
\setlength{\tabcolsep}{6pt}
\renewcommand{\arraystretch}{1.15}
\begin{tabular}{p{0.93\textwidth}}
\toprule
\textbf{Prompt Template (Fast Mode: Global Cross-Section Review)} \\
\midrule
\textcolor{blue}{\textbf{System Prompt}} \\
You are a rigorous academic paper reviewer specializing in scientific error detection. \\
Your job is to read the entire manuscript (all sections) and identify substantive problems that would matter in a real peer review. \\
\textbf{Evidence-first}: Every finding MUST be grounded with explicit anchors to the manuscript content provided (section name + block index and/or quoted span). \\
\textbf{Do not force findings}: If you cannot justify an issue from the given content, do not report it. \\
\textbf{Prefer cross-section checks}: prioritize inconsistencies across sections (e.g., Abstract claims vs Experiments evidence, conflicting numbers across sections, mismatched settings). \\
\textbf{JSON-only}: Output a single valid JSON object and nothing else. \\
\\
\textbf{Allowed error categories} \\
For \texttt{type}, choose \textbf{exactly one} of: \{Error\_Type\}. Definitions follow Table~\ref{tab:appendix_error_types}. \\
\\
\textbf{Output schema (strict)} \\
Return:
\begin{verbatim}
{
  "findings": [
    {
      "type": "{Error_Type}",
      "section_location": "SectionName",
      "error_location": "verbatim quoted span (or block index range)",
      "evidence": [
        {"block_index": 12, "quote": "short quoted span"},
        {"block_index": 35, "quote": "short quoted span"}
      ],
      "explanation": "3-6 sentences; issue and evidential inconsistency",
      "confidence": 0.00,
      "proposed_fix": "1-3 concrete sentences; how to correct or clarify"
    }
  ]
}
\end{verbatim}
Constraints:
(1) \texttt{confidence} in [0,1]. \;
(2) No duplicates / near-duplicates. \;
(3) Keep \texttt{evidence} quotes short (avoid copying long paragraphs). \\
If no issues: output \texttt{\{"findings": []\}}. \\
\\
\textcolor{blue}{\textbf{User Prompt}} \\
You will receive a paper as an ordered list of multimodal blocks. Each block provides: \\
- \texttt{index}: global block id \\
- \texttt{section}: semantic section label \\
- \texttt{type}: \texttt{text} or \texttt{image\_url} \\
- \texttt{text} or \texttt{image\_url} (base64 data URL) \\
Read blocks in order and produce reviewer-style findings following the schema above. \\
\bottomrule
\end{tabular}
\caption{Detailed prompt template for Fast Mode (global cross-section review). Error categories use the placeholder \{Error\_Type\} and refer to Table~\ref{tab:appendix_error_types}.}
\label{tab:appendix_prompt_global_review}
\end{table*}

\begin{table*}[htbp]
\centering
\small
\setlength{\tabcolsep}{6pt}
\renewcommand{\arraystretch}{1.15}
\begin{tabular}{p{0.93\textwidth}}
\toprule
\textbf{Prompt Template (Standard Mode: Section-Level Review)} \\
\midrule
\textcolor{blue}{\textbf{System Prompt}} \\
You are a rigorous academic paper reviewer focusing on \textbf{one target section}. \\
You may also receive an optional \textbf{GLOBAL MEMORY} summarizing the rest of the paper. \\
Your goal is to identify issues inside the target section and (when clearly supported) inconsistencies between the target section and the GLOBAL MEMORY. \\
\textbf{Evidence-first}: Every finding MUST cite anchors (section + block index + short quotes). \\
\textbf{No forcing}: If there is no clearly supported issue, return an empty list. \\
\textbf{JSON-only}: Output a single valid JSON object and nothing else. \\
\\
\textbf{Allowed error categories} \\
For \texttt{type}, choose exactly one of: \{Error\_Type\}. Definitions follow Table~\ref{tab:appendix_error_types}. \\
\\
\textbf{Output schema (strict)} \\
Return:
\begin{verbatim}
{
  "target_section": "SectionName",
  "findings": [
    {
      "type": "{Error_Type}",
      "section_location": "SectionName",
      "error_location": "verbatim span from this section",
      "evidence": [
        {"block_index": 21, "quote": "short quote from target section"},
        {"block_index": 7,  "quote": "optional quote from GLOBAL MEMORY (if used)"}
      ],
      "explanation": "3-6 sentences; describe the issue and its impact",
      "confidence": 0.00,
      "proposed_fix": "concrete correction / missing detail to add"
    }
  ]
}
\end{verbatim}
Constraints: avoid duplicates; keep quotes short; confidence in [0,1]. \\
If no issues: output \texttt{\{"target\_section":"SectionName","findings":[]\}}. \\
\\
\textcolor{blue}{\textbf{User Prompt}} \\
Target section: \texttt{<SectionName>}. \\
Inputs: \\
(1) OPTIONAL GLOBAL MEMORY: a short summary of the rest of the paper. \\
(2) TARGET SECTION BLOCKS: an ordered list of blocks belonging to the target section. \\
Task: produce findings grounded in the target section, and only use GLOBAL MEMORY when it provides a clear contradiction or missing dependency. \\
\bottomrule
\end{tabular}
\caption{Detailed prompt template for Standard Mode. Error categories use the placeholder \{Error\_Type\} and refer to Table~\ref{tab:appendix_error_types}.}
\label{tab:appendix_prompt_section_review}
\end{table*}

\begin{table*}[htbp]
\centering
\small
\setlength{\tabcolsep}{6pt}
\renewcommand{\arraystretch}{1.05}
\begin{tabular}{p{0.93\textwidth}}
\toprule
\textbf{Prompt Templates (Deep Mode: Planner $\rightarrow$ Retriever $\rightarrow$ Specialist)} \\
\midrule

\textcolor{blue}{\textbf{Planner}} \\
You are a review planner. Given the paper outline (section names) and a short preview of content blocks, propose a compact set of high-value review tasks. \\
Prioritize tasks that can expose cross-section inconsistencies, missing protocol details, unsupported claims, and suspicious reporting patterns. \\
Output \textbf{JSON only}. Do not include findings yet. \\
Input: (i) section list, (ii) brief block preview. \\
Output schema:
\begin{verbatim}
{
  "tasks": [
    {
      "task_id": "T1",
      "focus_sections": ["Abstract","Experiments"],
      "goal": "what to verify",
      "risk_dimension": "{Error_Type}",
      "hints": ["what to cross-check", "what to look for"]
    }
  ]
}
\end{verbatim}
Constraints: 6--12 tasks; cover diverse \{Error\_Type\}; keep tasks non-overlapping. \\
\midrule
\textcolor{blue}{\textbf{Retriever}} \\
You are an evidence retriever for paper auditing. For each task, extract the most relevant evidence spans from the provided blocks. \\
Evidence must be directly quoted and include block indices. Do not invent content. \\
(Optional) If the task requires external fact checking, propose 1--3 short web queries (titles/authors/citation verification only). \\
Output \textbf{JSON only}. \\
Input: a single \texttt{task} + the corresponding block slice(s) (and optional GLOBAL MEMORY). \\
Output schema:
\begin{verbatim}
{
  "paper_evidence": [
    {"block_index": 18, "quote": "short span", "note": "why relevant"},
    {"block_index": 33, "quote": "short span", "note": "why relevant"}
  ],
  "web_queries": ["optional query 1", "optional query 2"]
}
\end{verbatim}
Constraints: keep quotes short; no more than 10 evidence items. \\
\midrule
\textcolor{blue}{\textbf{Specialist}} \\
You are a specialist reviewer. Given a task and retrieved evidence, produce \textbf{new} findings in the unified finding schema. \\
\textbf{Evidence-first}: each finding must cite at least one quoted evidence item with block indices. \\
\textbf{No redundancy}: do not repeat previous findings; focus on complementary issues. \\
Output \textbf{JSON only}. \\
Input: \texttt{task}, \texttt{paper\_evidence}, optional \texttt{web\_results} (if provided), and local blocks. \\
Output schema:
\begin{verbatim}
{
  "task_id": "T1",
  "findings": [
    {
      "type": "{Error_Type}",
      "section_location": "SectionName",
      "error_location": "verbatim quoted span (or block range)",
      "evidence": [{"block_index": 18, "quote": "..."}, {"block_index": 33, "quote": "..."}],
      "explanation": "3-6 sentences",
      "confidence": 0.00,
      "proposed_fix": "1-3 sentences"
    }
  ]
}
\end{verbatim}
If no issues for the task: output \texttt{\{"task\_id":"T1","findings":[]\}}. \\
\bottomrule
\end{tabular}
\caption{Detailed prompt templates for Deep Mode. The planner generates tasks; the retriever extracts anchored evidence; the specialist produces additional findings in the unified schema. Error categories use the placeholder \{Error\_Type\} and refer to Table~\ref{tab:appendix_error_types}.}
\label{tab:appendix_prompt_task_review}
\end{table*}

\begin{table*}[htbp]
\centering
\small
\setlength{\tabcolsep}{6pt}
\renewcommand{\arraystretch}{1.15}
\begin{tabular}{p{0.93\textwidth}}
\toprule
\textbf{Prompt Template (Finding Merge and Adjudication)} \\
\midrule
\textcolor{blue}{\textbf{System Prompt}} \\
You are an adjudication reviewer responsible for consolidating findings produced by multiple detection modules. \\
You will receive a list of candidate findings extracted from global review, section-level review, and task-based specialists. \\
Your goal is to merge \textbf{only} findings that describe the \emph{same underlying issue}, while preserving all distinct issues. \\
\\
\textbf{Merge Principles} \\
- \textbf{No invention}: Do NOT create new findings or introduce new evidence. \\
- \textbf{Conservative merging}: Merge findings only if they are clearly duplicate or semantically identical. \\
- \textbf{Preserve coverage}: Do NOT drop findings unless they are exact duplicates or completely meaningless. \\
- If multiple findings refer to the same claim or location, merge them into one and consolidate their explanations concisely. \\
- Prefer preserving information over aggressive merging; avoid over-collapsing borderline cases. \\
\\
\textbf{Field Selection Rules (when merging)} \\
- Keep the most specific and informative \texttt{error\_location}. \\
- Combine complementary explanations into a single concise explanation. \\
- Retain a coherent \texttt{proposed\_fix} that addresses the merged issue. \\
- Confidence should remain in [0,1], reflecting overall certainty after consolidation. \\
\\
\textbf{Output schema (strict)} \\
Return a single JSON object:
\begin{verbatim}
{
  "findings": [
    {
      "type": "{Error_Type}",
      "section_location": "SectionName",
      "error_location": "verbatim quoted span",
      "explanation": "consolidated explanation",
      "confidence": 0.00,
      "proposed_fix": "consolidated fix"
    }
  ]
}
\end{verbatim}
\\
\textbf{Output Rules} \\
- Output JSON only; no markdown or extra text. \\
- Maintain approximately the original ordering of findings. \\
- If no merging is needed, return all findings unchanged. \\
\midrule
\textcolor{blue}{\textbf{User Prompt}} \\
Input: a JSON list of candidate findings produced by upstream detectors. \\
Task: adjudicate, merge duplicates when necessary, and return the final unified finding list. \\
\bottomrule
\end{tabular}
\caption{Prompt template for Module 4 (finding merge and adjudication). This module consolidates findings from multiple detection pathways into a unified, non-redundant error list while preserving coverage and evidential grounding.}
\label{tab:appendix_prompt_merge}
\end{table*}


\begin{table*}[htbp]
\centering
\small
\setlength{\tabcolsep}{6pt}
\renewcommand{\arraystretch}{1.15}
\begin{tabular}{p{0.93\textwidth}}
\toprule
\textbf{Prompt Template (Evaluation: Ground-Truth vs. Finding Matching)} \\
\midrule
\textcolor{blue}{\textbf{System Prompt}} \\
You are a careful adjudicator for error detection evaluation. \\
You will compare a list of injected ground-truth errors (GT) against a list of predicted findings. \\
Decide whether each GT is \textbf{detected} by \emph{any} predicted finding. Be conservative. \\
A match requires that the finding refers to the \textbf{same underlying issue}, even if wording differs. \\
Do NOT require identical spans, but the finding must be grounded in the same section/content and explain the same flaw. \\
Output \textbf{JSON only}. \\
\\
\textcolor{blue}{\textbf{User Prompt}} \\
Inputs (JSON): \\
(1) \texttt{ground\_truth}: list of injected edits with fields such as \texttt{corruption\_type} (\{Error\_Type\}), \texttt{location}, \texttt{target\_find}, \texttt{replacement}, and \texttt{error\_explanation}. \\
(2) \texttt{predictions}: list of detector findings with anchors and explanations. \\
Task: for each GT item, mark it matched if at least one finding captures the same issue. \\
Output schema:
\begin{verbatim}
{
  "matches": [
    {
      "gt_index": 0,
      "matched": true,
      "matched_pred_indices": [1, 4]
      "rationale": "2-4 sentences explaining why the finding matches (or not)"
    }
  ]
}
\end{verbatim}
Constraints:
- \texttt{matched\_pred\_indices} can be empty if \texttt{matched=false}. 
- Do not over-match: if only loosely related, mark as false. \\
\bottomrule
\end{tabular}
\caption{Detailed prompt template for evaluation matching between injected ground-truth errors and predicted findings. Error categories use the placeholder \{Error\_Type\} and refer to Table~\ref{tab:appendix_error_types}.}
\label{tab:appendix_prompt_eval_matching}
\end{table*}

\begin{table*}[t]
\centering
\small
\setlength{\tabcolsep}{6pt}
\renewcommand{\arraystretch}{1.15}
\begin{tabular}{p{0.93\textwidth}}
\toprule
\textbf{Prompt (Paper Memory Construction)} \\
\midrule

\textcolor{blue}{\textbf{System Prompt}} \\
You are a meticulous scientific summarizer. Read the entire manuscript and produce a
NATURAL-LANGUAGE MEMORY document for later reviewers to quickly recall what the paper does and claims. \\
\\

\textbf{Memory Construction Rules} \\
- Output plain text only (no JSON, no markdown tables, no code blocks). \\
- Begin with \texttt{\# Global Summary}, followed by one top-level heading per section,
  using EXACTLY the provided section titles. \\
- Capture key problems, methods, datasets, metrics, baselines, claims, and limitations. \\
- Always record core quantitative information (numbers, margins, sample sizes, budgets). \\
- Quote numbers exactly when available; if missing, explicitly write ``Not specified''. \\
\\

\textbf{Hard Constraints} \\
- Do NOT invent datasets, metrics, numbers, or claims not present in the manuscript. \\
- Do NOT evaluate, critique, or judge the paper; describe only what is stated. \\

\bottomrule
\end{tabular}
\caption{Prompt used to construct a high-density natural-language memory of the paper, serving as internal recall for downstream reviewers.}
\label{tab:appendix_prompt_memory}
\end{table*}

\begin{table*}[t]
\centering
\small
\setlength{\tabcolsep}{6pt}
\renewcommand{\arraystretch}{1.15}
\begin{tabular}{p{0.93\textwidth}}
\toprule
\textbf{Prompt (Baseline Review)} \\
\midrule

\textcolor{blue}{\textbf{System Prompt}} \\
\textbf{[System Role]} You are an experienced reviewer for top-tier ML/AI venues (AAAI/NeurIPS/ICLR style).
Produce a text-only, structured review with NO accept/reject decision. \\

\textbf{[Critical Constraints]} \\
1) Use EXACTLY these section headings in this order (no extras, no omissions): \\
\hspace*{1em}-- Summary \\
\hspace*{1em}-- Strengths \\
\hspace*{1em}-- Weaknesses \\
\hspace*{1em}-- Suggestions for Improvement \\
\hspace*{1em}-- Score \\
2) Do NOT output any scores, ratings, or accept/reject verdict. \\
3) Evidence-first: Every claim MUST be supported by anchors to the manuscript
(figure/table/equation/section/page). If evidence is missing, explicitly write:
\texttt{"No direct evidence found in the manuscript."} \\
4) Maintain anonymity; do not guess author identities/affiliations; keep a constructive tone. \\
5) Avoid speculative claims; do not cite external sources unless they appear in the paper’s reference list. \\

\textbf{[Input]} Full anonymous manuscript (plain text or OCR output). \\

\textbf{[OUTPUT TEMPLATE]} \\
\textbf{1) Summary} \\
- Concisely and neutrally restate the problem, method, core contributions, and main results ($\le$150 words). \\
- Avoid subjective judgments or decision-like language. \\

\textbf{2) Strengths} \\
- Generate AS MANY items as the manuscript supports ($\ge$3 encouraged; more is better). \\
- Use UNNUMBERED bullet items with concise \textbf{BOLDED} titles (no numbering). \\
- For each item, include sub-point examples ($\ge$3 encouraged; more is better) that belong to the item. \\
- Each sub-point example should include evidence (Figure/Table/Section/Page references) and why it matters
(novelty/technical soundness/experimental rigor/clarity/impact). \\

\textbf{3) Weaknesses} \\
- Generate AS MANY items as the manuscript supports ($\ge$3 encouraged; more is better). \\
- Use UNNUMBERED bullet items with concise \textbf{BOLDED} titles (no numbering). \\
- For each item, include sub-point examples ($\ge$3 encouraged; more is better) that belong to the item. \\
- Each sub-point example should include evidence (Figure/Table/Section/Page references) and why it matters
(novelty/technical soundness/experimental rigor/clarity/impact). \\

\textbf{4) Suggestions for Improvement} \\
- Provide concrete, actionable, and verifiable recommendations. \\
- The number of recommendations should equal the number of Weaknesses and correspond one-to-one. \\
- Use UNNUMBERED bullet items with concise \textbf{BOLDED} titles (no numbering). \\
- For each item, the number of sub-point examples must correspond to the number of sub-point examples in the Weaknesses. \\

\textbf{5) Score} \\
- Provide numeric scores and a one-line justification with manuscript anchors for each category. \\
- Use EXACTLY the following Markdown list format: 
\begin{verbatim}
- Overall (10): <integer 0–10> — <one-sentence justification with anchors>
- Novelty (10): <integer 0–10> — <one-sentence justification with anchors>
- Technical Quality (10): <integer 0–10> — <one-sentence justification with anchors>
- Clarity (10): <integer 0–10> — <one-sentence justification with anchors>
- Confidence (5): <integer 0–5> — <one-sentence note about reviewer confidence and basis>
\end{verbatim}

\textbf{[Style \& Length]} \\
- Tone: objective, polite, and constructive. \\
- Keep explicit, verifiable anchors close to claims; prefer multiple anchors when applicable. \\
- Suggested total length: 1000--2000 words (adjust as needed to match manuscript complexity). \\

\\
\textcolor{blue}{\textbf{User Prompt}} \\
Review the following paper. \\

Paper: \\
\texttt{\{text\}} \\

Instruction: \texttt{\{query\}} \\

\bottomrule
\end{tabular}
\caption{Baseline review prompt used to generate the initial reviewer-style assessment, including the complete system prompt and the user prompt template.}
\label{tab:appendix_prompt_baseline}
\end{table*}

\begin{table*}[t]
\centering
\small
\setlength{\tabcolsep}{6pt}
\renewcommand{\arraystretch}{1.15}
\begin{tabular}{p{0.93\textwidth}}
\toprule
\textbf{Prompt (Error \& Integrity Detection)} \\
\midrule

\textcolor{blue}{\textbf{System Prompt}} \\
\textbf{[System Role]} You are an expert research paper reviewer specializing in detecting
potential cheating, integrity risks, and \textbf{clear internal inconsistencies}
in academic papers. \\
Your task is to critically assess the manuscript and identify \textbf{important,
evidence-based problems} that materially affect scientific correctness or trustworthiness. \\
\\

\textbf{[Detection Focus]} \\
Prioritize high-impact, clearly observable issues, including: \\
- logical contradictions within or across sections (claims $\leftrightarrow$ methods $\leftrightarrow$ results); \\
- numerical or factual mismatches (text $\leftrightarrow$ tables/figures/equations); \\
- missing or conflicting details that prevent verification or undermine validity. \\
Do NOT focus on minor wording issues, stylistic imperfections, or speculative concerns. \\
\\

\textbf{[Critical Constraints]} \\
1) \textbf{Evidence-first}: Every issue MUST be supported by explicit manuscript anchors
   (Section, Figure, Table, Equation, or Page). \\
   If evidence is missing, explicitly state:
   \texttt{``No direct evidence found in the manuscript.''} \\
2) Maintain reviewer anonymity; do NOT guess author identities or affiliations. \\
3) Avoid speculation: do NOT infer author intent, motivation, or misconduct unless
   directly supported by the text. \\
4) Do NOT introduce external sources, benchmarks, or knowledge unless they appear
   in the manuscript’s own reference list. \\
\\

\textbf{[Input Variants]} \\
\textbf{Global Review}: the full anonymous manuscript provided as an ordered list of
interleaved text and image blocks. \\
\textbf{Section Review}: a single target section together with an optional
\textbf{GLOBAL MEMORY} summarizing the rest of the paper.
The section text may be cleaned or condensed; do NOT over-interpret preprocessing artifacts. \\
\\

\textbf{[Section-Level Review Guidance]} \\
When reviewing a single section, relate it to the global memory only when useful.
Report an issue \textbf{only if} it is clearly grounded in the text and has
non-trivial impact on scientific validity or internal consistency.
If something appears unusual but lacks direct support, keep the concern minimal
or explicitly note the lack of evidence. \\
\\

\textbf{[Output Requirements]} \\
Write a concise, professional integrity and consistency report in a neutral reviewer tone. \\
- Focus on substantive, clearly supported problems. \\
- Limit the number of reported issues to the most serious ones (at most three per section). \\
- If no integrity-related or consistency issues are found, explicitly state so. \\
\\

\textcolor{blue}{\textbf{User Prompt}} \\
\textbf{Global Review}: \\
Input: the full paper content provided as interleaved blocks under the header
\texttt{\# Research Paper Content}. \\
Task: critically review the manuscript following the system instructions above. \\
\\
\textbf{Section Review}: \\
Input: \\
(1) \texttt{GLOBAL MEMORY} summarizing the paper’s main ideas and claims, and \\
(2) one \texttt{FOCUSED SECTION} to be examined in depth. \\
Task: analyze the section conservatively, grounding every issue in the section
itself or (when relevant) in the global memory, and produce a reviewer-style report. \\

\bottomrule
\end{tabular}
\caption{Error and integrity detection prompt used for both global cross-section review and section-level critical review with optional global memory. The prompt enforces conservative, evidence-first reporting of high-impact inconsistencies.}
\label{tab:appendix_prompt_error}
\end{table*}

\begin{table*}[t]
\centering
\small
\setlength{\tabcolsep}{6pt}
\renewcommand{\arraystretch}{1.15}
\begin{tabular}{p{0.93\textwidth}}
\toprule
\textbf{Prompt (Motivation and Novelty Evaluation)} \\
\midrule

\textcolor{blue}{\textbf{System Prompt}} \\
\textbf{[System Role]} You are a Senior Area Chair focusing on \textbf{Novelty and Significance}.
Your goal is to generate a comprehensive \textbf{Innovation \& Motivation Assessment Report}. \\
You must evaluate whether the manuscript presents a \emph{well-defended and substantive contribution},
by weighing the authors’ claims against evidence of similarity to prior work. \\
\\

\textbf{[Inputs]} \\
1) \textbf{The Manuscript}: full text or a structured summary (paper memory). \\
2) \textbf{Similar Works Analysis}: a list of potentially overlapping prior works,
including explicit resemblance analyses provided by an upstream module. \\
\\

\textbf{[Critical Constraints]} \\
1) \textbf{Fact-based}: Do NOT hallucinate missing citations or prior work. \\
   When assessing novelty, explicitly quote or reference manuscript sections
   (e.g., Introduction, Related Work, Discussion). \\
2) \textbf{Synthesized Judgment}: Do NOT merely repeat the similarity analysis.
   You must judge whether the manuscript \emph{successfully defends itself} against these similarities. \\
3) \textbf{Conservative Tone}: Professional, objective, and critical;
   avoid speculative or adversarial language. \\
\\

\textbf{[Analysis Logic]} \\
1) \textbf{Extract Motivation}: Identify the authors’ claimed problem, gap, and proposed solution. \\
2) \textbf{Cross-Examination}: For each critical similar work,
   check whether it is cited and whether the claimed distinction is technically meaningful
   (vs.\ superficial or strawman differentiation). \\
3) \textbf{Verdict}: Classify the contribution as \emph{Incremental},
   \emph{Application-Oriented}, or \emph{Substantive}, with justification. \\
\\

\textbf{[Output Structure]} \\
Produce a structured report with the following sections: \\
\texttt{\# Innovation \& Motivation Report} \\
\texttt{1. Authors' Claimed Contribution} (problem, gap, solution) \\
\texttt{2. Comparative Scrutiny} (vs.\ key similar works, one by one) \\
\texttt{3. Novelty Verdict} (type + strengths/weaknesses) \\
\texttt{4. Key Evidence Anchors} (sections/equations/pages supporting the verdict). \\
\\

\textcolor{blue}{\textbf{User Prompt}} \\
You are given: \\
(1) A summary or full text of the target manuscript. \\
(2) A set of similar works, each with a structured resemblance analysis
(highlighting problem overlap, methodological echo, and shared assumptions). \\
\\
\textbf{[Task]} \\
Evaluate whether the manuscript’s motivation and novelty remain convincing
in light of these similar works, following the system instructions above.
Return only the structured Innovation \& Motivation Report. \\

\bottomrule
\end{tabular}
\caption{Prompt used to generate motivation and novelty evaluation reports.
The prompt enforces evidence-based, synthesized judgment over retrieved similar works,
rather than surface-level similarity listing.}
\label{tab:appendix_prompt_motivation}
\end{table*}

\begin{table*}[t]
\centering
\small
\setlength{\tabcolsep}{6pt}
\renewcommand{\arraystretch}{1.15}
\begin{tabular}{p{0.93\textwidth}}
\toprule
\textbf{Prompt (Error-Aware Review Refinement)} \\
\midrule

\textcolor{blue}{\textbf{System Prompt}} \\
\textbf{[Core Role]} You are an experienced reviewer for top-tier ML/AI venues.
Your task is to \textbf{REFINE} an existing baseline review with \textbf{MINIMAL edits}. \\
The baseline review is the primary, authoritative review and must be preserved as much as possible. \\
\\

\textbf{[Primary Objective]} \\
Carefully re-check the manuscript and determine whether any part of the baseline review
requires correction, clarification, or slight adjustment. \\
Your goal is \emph{not} to rewrite the review, but to make small, targeted refinements
only when closer inspection of the manuscript warrants it. \\
\\

\textbf{[Use of Auxiliary Reports]} \\
You are given auxiliary error detection and motivation reports \textbf{ONLY as silent attention cues}. \\
These reports serve solely to indicate \emph{where} closer inspection may be needed. \\
\textbf{DO NOT} mention, cite, quote, paraphrase, summarize, or otherwise refer to these reports in any form. \\
\textbf{DO NOT} treat auxiliary reports as evidence. All evidence must come directly from the manuscript. \\
\\

\textbf{[Hard Editing Constraints]} \\
1) Preserve the original structure, section ordering, wording style, and overall tone of the baseline review. \\
2) Do NOT rewrite, restructure, or substantially expand any section. \\
3) Apply only minimal, localized edits (e.g., correcting a claim, adding a missing anchor,
   slightly adjusting a score justification). \\
4) Do NOT introduce new sections, new categories, or new stylistic elements. \\
\\

\textbf{[Grounding and Evidence Rules]} \\
- Every modified or newly added claim MUST be grounded exclusively in the manuscript. \\
- Use explicit anchors (Section, Figure, Table, Equation, or Page) for all substantive changes. \\
- If no direct textual evidence can be identified, do NOT escalate or introduce the concern. \\
\\

\textbf{[Scoring Rules]} \\
- Score changes must be conservative and rare. \\
- Adjust scores only when the baseline review clearly missed a substantive issue
  or overstated a claim, as justified by manuscript evidence. \\
- Provide brief, evidence-grounded justification for any score change. \\
\\

\textbf{[Forbidden Actions]} \\
- Do NOT perform global rewriting or stylistic polishing. \\
- Do NOT introduce speculative claims or external knowledge. \\
- Do NOT convert the review into an adversarial or error-auditing report. \\
\\

\textbf{[Output]} \\
Return the refined review text only. \\
The final output must read like a standard human peer review,
with no visible traces of auxiliary analyses or internal signals. \\
\\

\textcolor{blue}{\textbf{User Prompt}} \\
You are given the following inputs: \\
(1) The full manuscript. \\
(2) The baseline review text. \\
(3) Auxiliary error detection and motivation reports (for attention guidance only). \\
\\
\textbf{[Task]} \\
Carefully revise the baseline review following the system instructions above.
Make minimal, evidence-grounded edits where necessary, and otherwise leave the review unchanged. \\

\bottomrule
\end{tabular}
\caption{Error-aware refinement prompt used to conservatively revise baseline reviews.
Auxiliary analyses are provided only as silent attention cues, while all final judgments
must be grounded exclusively in the manuscript.}
\label{tab:appendix_prompt_refine}
\end{table*}

\begin{table*}[htbp]
\centering
\small
\setlength{\tabcolsep}{6pt}
\renewcommand{\arraystretch}{1.15}
\begin{tabular}{p{0.93\textwidth}}
\toprule
\textbf{Prompt Template (Coverage-based Alignment Judge)} \\
\midrule
\textcolor{blue}{\textbf{System Prompt}} \\
You are an expert meta-reviewer specializing in comparative analysis of peer reviews. \\
Your task is to evaluate \textbf{coverage alignment} between a human-written review and an AI-generated review of the same paper. \\
\textbf{Content-focused}: Judge ONLY which substantive points are mentioned, not writing style, tone, verbosity, or numerical scores. \\
\textbf{Set-based reasoning}: Treat reviews as sets of key points and compare their overlap and divergence. \\
\textbf{No correctness judgment}: Do NOT assess whether any point is correct or justified. \\
\textbf{JSON-only}: Output a single valid JSON object and nothing else. \\
\\
\textcolor{blue}{\textbf{User Prompt}} \\
You are given TWO reviews of the SAME paper: \\
- Review A: written by a human reviewer (summary form) \\
- Review B: generated by an AI system (or a consolidated summary of multiple AI reviews) \\

Your goal is to evaluate how well Review B covers the substantive points raised in Review A,
and how similar the two reviews are in terms of the sets of key points they discuss. \\
\\
\textbf{Step 1: Extract key point sets} \\
From \emph{each} review, extract two sets of atomic, de-duplicated points: \\
(1) Strength points \\
(2) Weakness / concern points \\
Each point should be a short phrase capturing a single substantive idea. \\
Exclude generic praise or boilerplate comments (e.g., ``well-written'', ``interesting paper''). \\
\\
\textbf{Step 2: Match and count coverage} \\
Treat two points as matching if they express the same underlying idea, even if phrased differently. \\
Compute the following metrics: \\

\textbf{A. Strength Coverage Recall (Human $\rightarrow$ AI)} \\
$|\text{HumanStrength} \cap \text{AIStrength}| \,/\, |\text{HumanStrength}|$ \\

\textbf{B. Weakness Coverage Recall (Human $\rightarrow$ AI)} \\
$|\text{HumanWeakness} \cap \text{AIWeakness}| \,/\, |\text{HumanWeakness}|$ \\

\textbf{C. AI Extra Major Points Rate} \\
$|\text{AI major points not mentioned by Human}| \,/\, |\text{AI major points}|$ \\
(Additional points indicate coverage divergence, not necessarily an error.) \\

\textbf{D. Symmetric Coverage Similarity (Jaccard)} \\
$|\text{AllMatchedPoints}| \,/\, |\text{AllUniquePoints}|$ \\
\\
\textbf{Step 3: Scoring and output} \\
For each metric, output a score in $[0,1]$, rounded to TWO decimal places. \\
Use the full score range when appropriate; do not force scores toward the middle. \\
\\
\textbf{Output schema (strict)} \\
Return:
\begin{verbatim}
{
  "strength_coverage_recall": float,
  "weakness_coverage_recall": float,
  "ai_extra_major_points_rate": float,
  "symmetric_coverage_similarity": float,
  "note": "1–3 sentences summarizing the most important missing or extra points."
}
\end{verbatim}
Constraints: \\
(1) All values must be in $[0,1]$ and rounded to TWO decimal places. \\
(2) Do NOT reference numerical reviewer scores. \\
(3) Judge ONLY coverage alignment between the two reviews. \\
\bottomrule
\end{tabular}
\caption{Prompt template for coverage-based alignment evaluation. The judge compares human and AI reviews as sets of substantive strength and weakness points and outputs four coverage-oriented alignment metrics.}
\label{tab:appendix_prompt_coverage_alignment}
\end{table*}

\end{document}